\documentclass[10pt,letterpaper]{article}

\usepackage{cogsci}

\cogscifinalcopy 

\usepackage{booktabs}

\usepackage{graphicx} 
\usepackage{pslatex}
\usepackage{apacite}
\usepackage{graphicx}  
\usepackage{verbatim}
\usepackage{xcolor}
\usepackage[colorinlistoftodos]{todonotes}
\usepackage{natbib}
\usepackage{amsmath}
\usepackage{color}
\usepackage[most]{tcolorbox}
\usepackage{float} 
\usepackage{listings}
\definecolor{githubblue}{RGB}{49,46,138}
\definecolor{schemegreen}{RGB}{15,137,15}
\lstdefinelanguage{Scheme}{
  morekeywords=[1]{define, define-syntax, define-macro, lambda, define-stream, stream-lambda},
  morekeywords=[2]{begin, call-with-current-continuation, call/cc,
    call-with-input-file, call-with-output-file, case, cond, condition,
    do, else, for-each, if,
    let*, let, let-syntax, letrec, letrec-syntax,
    let-values, let*-values,
    and, or, not, delay, force,
    quasiquote, quote, unquote, unquote-splicing,
    map, fold, syntax, syntax-rules, eval, environment, query },
  morekeywords=[3]{import, export},
  alsodigit=!\$\%&*+-./:<=>?@^_~,
  sensitive=true,
  morecomment=[l]{;},
  morecomment=[s]{\#|}{|\#},
  morestring=[b]",
  basicstyle=\small\ttfamily,
  keywordstyle=\bf\ttfamily\color[rgb]{0,.3,.7},
  commentstyle={\color[rgb]{0.24, 0.51, 0.51}},
  stringstyle={\color[rgb]{0.75, 0.49, 0.07}},
  upquote=true,
  breaklines=true,
  breakatwhitespace=true,
  literate=*{`}{{`}}{1},
  showstringspaces=false
}
\lstdefinestyle{churchstyle}{
  backgroundcolor=\color{white},   commentstyle=\color{gray},
  keywordstyle=\color{githubblue},
  numberstyle=\color{black}\tiny,
  stringstyle=\color{red},
  basicstyle=\ttfamily\color{githubblue},
  breakatwhitespace=false,         
  breaklines=true,                 
  captionpos=b,                    
  keepspaces=true,                 
  numbers=none,                    
  numbersep=5pt,                  
  showspaces=false,                
  showstringspaces=false,
  showtabs=false,                  
  tabsize=2,
  literate=*{\{}{{\textcolor{NavyBlue}{\{}}}{1}
        {\}}{{\textcolor{black}{\}}}}{1}
        {[}{{\textcolor{black}{[}}}{1}
        {]}{{\textcolor{black}{]}}}{1}
        {(}{{\textcolor{black}{(}}}{1}
        {)}{{\textcolor{black}{)}}}{1}%
}
\usepackage{xcolor}
\usepackage{hyperref}
\hypersetup{
    colorlinks,
    linkcolor={red!50!black},
    citecolor={blue!50!black},
    urlcolor={blue!80!black}
}

\usepackage{amsfonts}

\definecolor{asparagus}{rgb}{0.53, 0.66, 0.42}

\global\hyphenpenalty=1000

\renewcommand{\thefootnote}{$\dagger$}
\newcommand\blfootnote[1]{%
  \begingroup
  \renewcommand\thefootnote{}\footnote{#1}%
  \addtocounter{footnote}{-1}%
  \endgroup
}

\usepackage{listings}
\usepackage{xcolor}
\definecolor{commentcolor}{RGB}{128, 128, 128}
\definecolor{keywordcolor}{RGB}{127, 0, 85}
\definecolor{functioncolor}{RGB}{0, 0, 255}
\lstdefinestyle{customjs}{
    language=JavaScript,
    basicstyle=\ttfamily\small,
    breaklines=true,
    columns=flexible,
    commentstyle=\itshape\color{commentcolor},
    keywordstyle=\color{keywordcolor},
    stringstyle=\color{functioncolor},
    numbers=left,
    numberstyle=\tiny\color{commentcolor},
    numbersep=10pt,
    tabsize=2,
    showstringspaces=false,
    captionpos=b,
    frame=single,
    morecomment=[l]{//},
    morekeywords={var, function, return},
    literate={*}{}{0\discretionary{}}
}

\title{Modeling Open-World Cognition as On-Demand Synthesis of Probabilistic Models}

 \author{ \\ 
 {
 \large\bf Lionel Wong*\textsuperscript{1,2}}\footnote{Corresponding email: \texttt{liowong@stanford.edu}. Appeared in the Proceedings of the Cognitive Science Conference. * indicates equal first-author contribution. + indicates equal senior author contribution.}, \quad 
 {\large\bf Katherine M. Collins*\textsuperscript{2,3}}, \quad 
 {\large\bf Lance Ying\textsuperscript{2,4}}, \quad 
{\large\bf Cedegao E. Zhang\textsuperscript{2}}, \\ 
{\large\bf Adrian Weller\textsuperscript{3}},  \quad 
{\large\bf Tobias Gerstenberg\textsuperscript{1}}, \quad 
{\large\bf Timothy O'Donnell\textsuperscript{5}},\quad 
{\large\bf Alexander K. Lew\textsuperscript{2,6}}, \\
{\large\bf Jacob D. Andreas\textsuperscript{2}},\quad 
{\large\bf Joshua B. Tenenbaum+\textsuperscript{2}},\quad 
{\large\bf Tyler Brooke-Wilson+\textsuperscript{6}} \\ \\
\textsuperscript{1}Stanford University \quad \textsuperscript{2}MIT \quad \textsuperscript{3}University of Cambridge \quad \textsuperscript{4}Harvard University \quad \textsuperscript{5}McGill University \quad \textsuperscript{6}Yale University \\
Correspondence to \texttt{liowong@stanford.edu}, \, \texttt{katiemc@mit.edu}, \, \texttt{tyler.brooke.wilson@yale.edu}
 }

\begin{document}

\maketitle

\begin{abstract}
When faced with novel situations, people are able to marshal relevant considerations from a wide range of background knowledge and put these to use in inferences and predictions. What permits us to draw in globally relevant information and reason over it coherently? Here, we explore the hypothesis that people use a combination of distributed and symbolic representations to construct bespoke mental models tailored to novel situations. We propose a computational implementation of this idea -- a ``Model Synthesis Architecture'' (MSA) -- using language models to implement global relevance-based retrieval and model synthesis and probabilistic programs to implement bespoke, coherent world models. We evaluate our MSA as a model of human judgments on a novel reasoning dataset. The dataset -- built around a `Model Olympics` domain of sports vignettes -- tests models' capacity for human-like, open-ended reasoning by requiring (i) judgments about novel causal structures described in language; (ii) drawing on large bodies of background knowledge; and (iii) doing both in light of observations that introduce arbitrary novel variables. Our MSA approach captures human judgments better than language model-only baselines, under both direct and chain-of-thought generations from the LM that supports model synthesis. These results suggest that MSAs can be implemented in a way that mirrors people's ability to deliver locally coherent reasoning over globally relevant variables, offering a path to understanding and replicating human reasoning in open-ended domains. 


\end{abstract}

\section{Introduction}

An influential idea in cognitive science holds that people reason and plan using mental models, or structured internal representations that mirror aspects of the world \citep{craik1943nature,johnson1980mental,gentner1983electricity}. In this view, people draw on structured mental models to maintain consistent beliefs about current world states, integrate new information into their beliefs, and evaluate the plausibility of alternative hypotheses or possible futures. 

\blfootnote{*Indicates equal first author contribution. + indicates equal senior author contribution. Presented at CogSci 2025.}

\begin{figure}[t]  
    \flushright  
    \includegraphics[width=\linewidth]{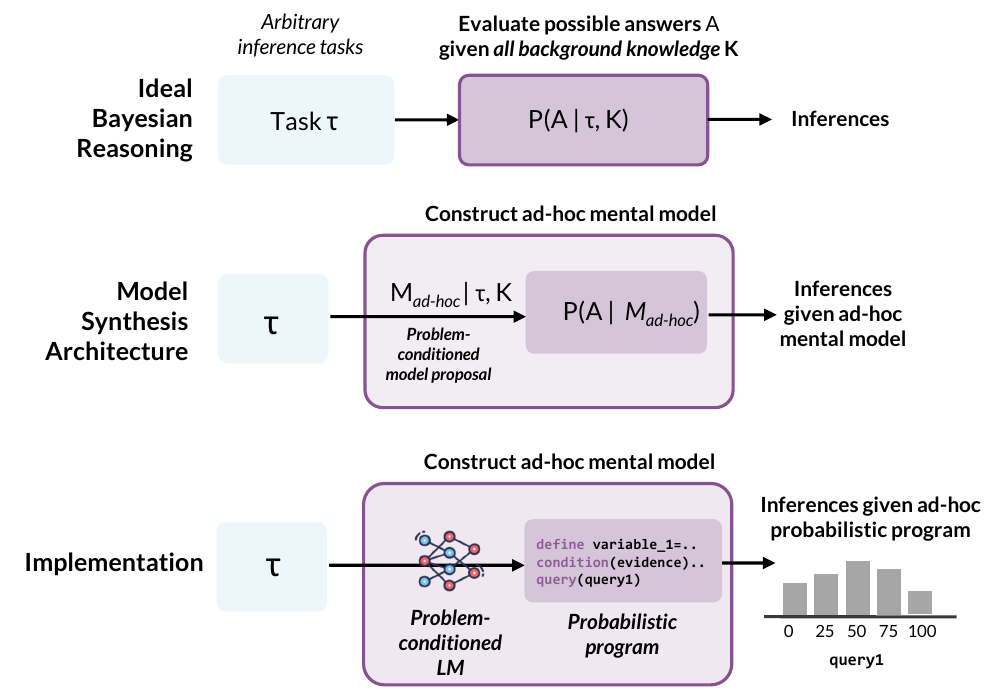}
    \caption{(Top) Idealized, coherent reasoning about arbitrary situations and reasoning tasks $\tau$ given all potentially relevant global beliefs and background knowledge $K$ raises problems of computational tractability. (Middle) We hypothesize that minds implement \textit{Model Synthesis Architectures}, using global relevance functions to  constructing ad-hoc symbolic mental models which permit locally coherent reasoning over select variables. (Bottom) We implement an MSA using LMs to parameterize global relevance functions and Probabilistic Programs to construct arbitrary probabilistic models.}
    \label{fig:marshaling-model}
    \vspace{-0.5cm}
\end{figure}

Bayesian modeling in cognitive science~\citep{griffithsbayesian, tenenbaum2011grow} has found significant empirical support for a version of this idea, showing that human judgments across a wide variety of tasks are well-modeled by inference and decisions in causal, probabilistic models. These models are coherent by design – representing the dependencies between relevant considerations, allowing for the integration of novel observations and drawing consistent predictions. Models of this kind have been successful in reproducing human judgments across domains as diverse as physical prediction \citep{battaglia2013simulation, hamrick2015think}, causal reasoning \citep{gerstenberg2021csm,beller2025language}, causal learning \citep{gopnik2004theory}, inferring people's beliefs \citep{baker2017rational}, desires \citep{jara2016naive}, goals and plans \citep{zhi2020online,baker2009action}, and communicative intentions \citep{frank2012predicting}. They have been successfully applied to reproducing human learning from sparse data in domains such as video games \citep{tsividis2021human}, word learning \citep{xu2007word}, or simple visual concepts \citep{lake2015human}. 

While these models are predictive of human judgments and learning in each of these settings, they remain importantly limited in that each model operates \textit{only} in the limited scope for which it was designed. Any given model can provide inferences over the variables it represents, but cannot handle novel considerations that were not part of the initial model specification. People, in contrast, regularly reason about novel questions (the cause and duration of an unusual traffic pattern, the emotional state of a friend or colleague at lunch, the odds of our favorite team winning the basketball game when their star player is out) that draw on a highly varied set of things we know about the world (our beliefs about weather, holidays, drivers, cultural norms, meal prices, athletes, and the rules of basketball, to name just a few), any of which are likely to be missing from any given mental model. To date, Bayesian modeling has left it unclear how such modeling approaches could scale to explain the simultaneous flexibility and coherence of human reasoning in general. 

So, how do people reason in \textit{locally coherent} ways in any given context, while drawing \textit{globally} on potentially relevant considerations across their background knowledge and beliefs? In this paper, we explore the hypothesis that human minds implement ``Model Synthesis Architectures” (MSAs), or architectures that construct small, ad-hoc mental models on the fly in response to task demands \citep{brooke-wilson2023bounded}. By reasoning within small models, MSAs can deliver local coherence over the variables they explicitly represent, while the ability to synthesize arbitrary models as needed allows the architecture to reason and plan in open-ended environments, where the relevant considerations are not fixed in advance. 

MSAs approach the problem of modeling human open-world reasoning by breaking it into two subproblems: (1)  constructing, or 'synthesizing,' ad-hoc models, which must include the relevant variables on a given occasion, and (2) reasoning within a model, which can be implemented via general algorithms for belief updating and decision making that have long been studied in cognitive science. Here we explore the MSA hypothesis in the setting of reasoning about causal systems from natural language inputs. This is a common, naturalistic setting that is intrinsically open-world, in that there is no natural limit on the kinds of variables and dependencies that could be relevant.  

We implement a concrete instance of an MSA using Probabilistic Programming Languages (PPLs) \citep{goodman2012church,bingham2019pyro,carpenter2017stan,cusumano2019gen} to express individual models as probabilistic programs, and using a neurally-guided program synthesis procedure, constituted by structured calls to a Language Model (LM), to construct relevant mental models. This represents one concrete implementation of an MSA. MSAs implemented in this way have desirable generality properties, as they are able to operate over arbitrary natural language inputs, because of their LM front end, while being able to express arbitrary probabilistic models, due to their general-purpose PPL modeling language. These choices correspond to two cognitive hypotheses about how these subproblems are accomplished in the human mind – first (i), that surfacing and organizing relevant background knowledge into mental models draws on large-scale statistical representations relating features of a problem, including especially linguistic features, and a compositional, code-like Language of Thought, as language-to-code LM pipelines can. And second (ii), that coherent reasoning and decision-making for novel problems is, at times, realized by principled, probabilistic algorithms for belief updating and decision making that run over mental models expressed in this language. 

Viewed from the side of Bayesian modeling, the goal in combining these is to build a system that, like human cognition, can operate in the open-world setting while still delivering the natively coherent reasoning of probabilistic models. Viewed from the side of LM-based modeling, the goal is to deliver a system that has the flexibility of LMs, a feature that makes them a compelling cognitive architecture in their own right \citep{binz2024centaur, carvalho2025naturalistic}, while addressing the fragility of their own internal ``world model''-like representations (e.g., \citeauthor{vafa2024evaluating}\citeyear{vafa2024evaluating}) especially in supporting robust and coherent reasoning on problems significantly outside of model training distributions \citep{mccoy2023embers}. 

We study this MSA implementation empirically and compare it with human ad-hoc reasoning as well as pure LM and PPL baselines using a domain of natural language inference tasks designed to test generalization and coherence. Our ``Model Olympics” domain consists of natural language vignettes about sporting events featuring distinct causal structures, unseen variables, and arbitrary considerations contributed by naive human participants. This domain naturally integrates intuitive causal reasoning, uncertainty, and diverse latent variables,  building on and generalizing established cognitive modeling \citep{gerstenberg2012pingpong} while providing a structured yet open-ended setting for evaluating flexible cognitive modeling architectures.  One cannot know, in advance of encountering a new sport scenario, whether injuries, weather, team dynamics, new strategies or equipment, or any of an unbounded set of potentially relevant variables will in fact be relevant. Such an open-world setting outstrips what can be handled by traditionally static Bayesian models. 

We design three experiments with human participants in this domain, with successive experiments demanding progressively more generalization to novel language, dependencies, and variables, while drawing on more distant background knowledge to do so. \textbf{Experiment 1} asks people to reason about three different sports of varying novelty (tug-of-war, canoe racing, and biathlon), and presents detailed vignettes that must be integrated to draw inferences about athletes (eg. their \textit{strength} or \textit{shooting accuracy}) and predict upcoming outcomes (e.g., \textit{who would win and by how much}?). \textbf{Experiment 2 }asks people to draw these same inferences and predictions from less detailed vignettes, which identify the relevant variables but require people to surface the causal relationships between them from background knowledge. \textbf{Experiment 3} introduces novel variables and dependencies into each vignette which then must be integrated into reasoning. These are further natural language 'observations' sourced from naive human subjects (instructed to provide novel considerations that would 'change their predictions about new matches'). As such, \textbf{Experiment 3} directly tests reasoning in an explicitly open-world setting.

Across all experiments, we find that human reasoning is well-captured by our Model Synthesis Architecture, which provides a better match to human judgments than LM-only baselines and model ablations. This represents a proof of concept that neural language modeling and structured probabilistic modeling can be interleaved to explain people's ability to reason in ways that are globally relevant and locally coherent in an open-world setting.

\begin{figure*}[t!]
    \centering
    \includegraphics[width=\linewidth]{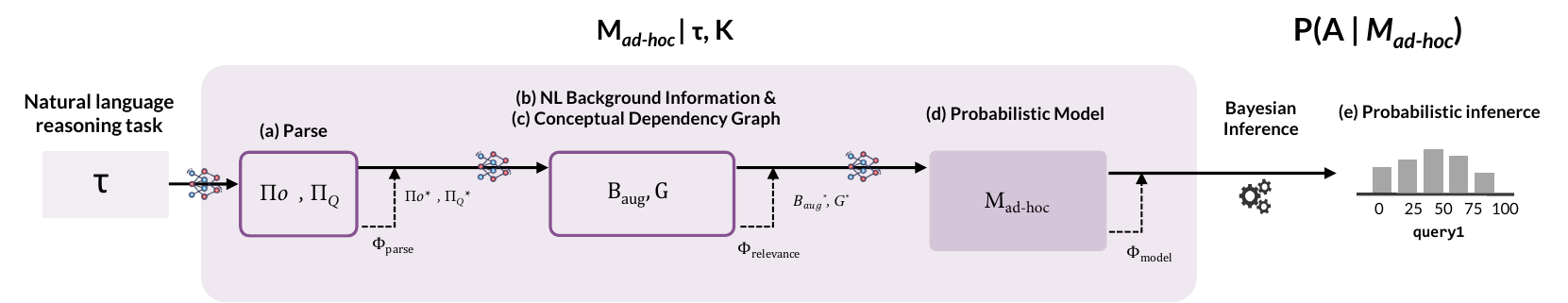}
    \vspace{-0.7cm}
    \caption{Schematic overview of our MSA implementation, which sequentially constructs $M_\text{ad-hoc}$ from input natural language tasks $\tau$ through interleaved LM-guided generation steps and scoring steps using intermediate scoring functions $\Phi$.}
    \label{fig:marshaling-model-schematic}
\end{figure*}
\begin{figure*}[t!]
    \centering
    \includegraphics[width=\linewidth]{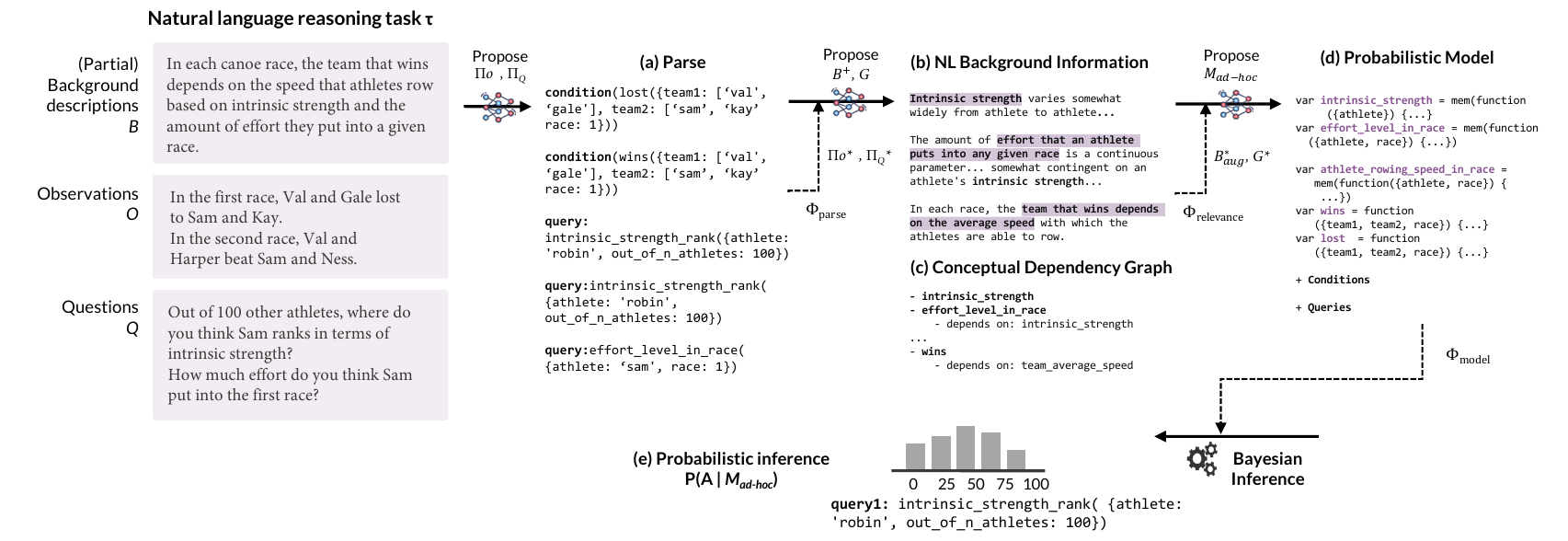}
    \vspace{-0.7cm}
    \caption{Detailed overview of our MSA implementation. Given input tasks $\tau$ as (potentially impoverished) background descriptions $B$, concrete observations $O$, and questions $Q$, we sequentially construct  $M_\text{ad-hoc}$ by: \textbf{(a)} \textit{parsing} the inputs into candidate probabilistic program expressions $\Pi_O, \Pi_Q$ for the observations and questions; \textbf{(b)} retrieving additional \textit{natural language background knowledge} $B^+$ to create an augmented background $B_{aug}$ that contains detailed, fully specified descriptions of intermediate latent variables and causal dependencies; and simultaneously \textbf{(c)} proposing a \textit{conceptual dependency graph} $G$ between proposed background variables; and then finally \textbf{(d)} synthesizing a formal symbolic \textit{probabilistic model}  $M_\text{ad-hoc}$ as a probabilistic program. We then compute \textbf{(e)}  \textit{model-based Bayesian inferences} to answer the questions given the synthesized probabilistic program. At each stage, candidate generations proposed by an LM are scored under utility functions $\Phi$, and the best scoring candidates (marked with a *) are passed on to the next stage of the model synthesis pipeline.}
    \label{fig:marshaling-model-detailed}
\end{figure*}

\section{Model Synthesis Architectures} 
We consider the general problem of inferring answers $A$ to an arbitrary inference or prediction problem $\tau$. In the idealized Bayesian inference setting (\autoref{fig:marshaling-model}, top), drawing these inferences involves conditioning on the information in the specific problem $\tau$ in light of \textit{all of the reasoner's prior background knowledge} $K$ to produce answers:
\setlength{\abovedisplayskip}{4pt}
\setlength{\belowdisplayskip}{4pt}
\begin{equation}
    P(A \mid \tau, K).
\end{equation}
Computing this will often be prohibitively costly, as probabilistic inference is intractable in the general case, and wasteful, since on any one occasion, it's likely that only a small portion of what the reasoner actually knows will matter for the question at hand. Instead of computing this full conditional, we propose that a reasoner ``marshals'' only a subset of their background knowledge ($K'$) that is relevant to the problem at hand, such that:
\setlength{\abovedisplayskip}{4pt}
\setlength{\belowdisplayskip}{4pt}
\begin{equation}
  P(A | \tau, K') \approx P(A \mid \tau, K).
\end{equation}
In particular, to draw coherent inferences, we propose that reasoners use this reduced set of background knowledge to construct a \textit{context-specific mental model} ($M_\text{ad-hoc}$) which they use to perform probabilistic inferences (\autoref{fig:marshaling-model}, middle), assuming that:
\setlength{\abovedisplayskip}{4pt}
\setlength{\belowdisplayskip}{4pt}
\begin{equation}
  P(A | M_\text{ad-hoc}) \approx P(A | \tau, K').
\end{equation}

We call any system that implements this overarching cognitive hypothesis a \textit{Model Synthesis Architecture}, as it decomposes reasoning about an arbitrary problem into two distinct computational subproblems: (1) a problem-conditioned, \textit{ad-hoc model synthesis} step to construct $M_\text{ad-hoc}$, and then (2) a step computing \textit{Bayesian model-based inferences} to answer the question conditional on the constructed model, i.e. computing $P(A | M_\text{ad-hoc})$. 

The formal nature of each of these subproblems differs. As in  \textit{resource-rational} framings \citep{lieder2020resource}, we propose that reasoners treat ad-hoc model synthesis as an optimization problem, selecting representations that they believe will be \textit{useful} for reasoning about a problem: 
\setlength{\abovedisplayskip}{4pt}
\setlength{\belowdisplayskip}{4pt}
\begin{equation}
 \text{argmax}_{i \in k_{model}}\Phi(M^{i}_\text{ad-hoc}, \tau)
\label{eq:useful}
\end{equation}
based on various model evaluation functions $\Phi$ (e.g., trading off between computational costs of inference in the model with expected accuracy for a set of queries) over a set of $k_{model}$ sampled models. 

In contrast to model synthesis, reasoning and planning with a synthesized model might look like optimization, inference, or deduction, depending on the problem and synthesized model. We focus on probabilistic inference, where models represent structured priors or \textit{conceptions} of the relevant variables and dependencies for the problem at hand, cf. \citep{gerstenberg2017intuitive,goodman2014concepts}. 

In this paper, we consider a subset of $\tau$, the space of natural language probabilistic reasoning problems defined by a tuple  $(B, O, Q)$, where $B$ is a (potentially partial and underspecified) set of \textit{background} variable descriptions $b_1, ..., b_N$ about a situation at hand (e.g., someone trying to predict upcoming sports tournaments in a bracket might mention factors like injuries or training that they believe should be considered);  $O$ is a set of \textit{observations} $o_1, ..., o_N$ providing evidence that bears on those variables (e.g., observations about which teams have previously won or lost in the tournament); and $Q$ is a set of \textit{questions} $q_1, ..., q_N$ that single out particular queries to answer given the evidence (e.g., specific prediction questions about which teams will win in an upcoming match). 

\subsection{Representing and synthesizing ad-hoc models}
The general MSA hypothesis breaks down reasoning into model synthesis and reasoning within a model. In this section, we describe a concrete MSA implementation in which ad-hoc models are represented as task-specific \textit{probabilistic programs}. Each probabilistic program represents models as a tuple $M_\text{ad-hoc} = (\Pi_{B}, \Pi_O, \Pi_Q)$, in which $\Pi_{B}$ are set of stochastic function definitions that formalize a causal prior over relevant background variables, by defining their distributional form and causal dependencies; $\Pi_O$ are a set of observed constraints over defined variables which condition belief under the prior; and $\Pi_Q$ are query expressions over the defined variables which define targets for Bayesian inference under the  conditioned probabilistic model. 

Our concrete implementation then frames model synthesis as \textit{LM-guided probabilistic program synthesis}, using LMs to parameterize a search procedure over programs conditioned on an input task, and to parameterize a set of model evaluation functions $\Phi$. This implementation ultimately answers queries in synthesized models as $P(A | M_\text{ad-hoc})$, using automatic Bayesian inference procedures defined generally over the probabilistic programming language. 

Our overarching goal is to synthesize \textit{useful} models for tasks (\autoref{eq:useful}). We present an implementation that approximates this optimization over models via a \textit{sequentially staged} synthesis process, interleaving structured steps of partial model generation and evaluation. Interleaving generation and evaluation allows us to focus future generation stages on outputs that evaluate highly under components of $\Phi$ so far, providing efficiency gains. This implementation sequentially constructs $M_\text{ad-hoc}$ by:
\vspace{2em}

\begin{itemize}
    \item \textbf{Parse:} (\autoref{fig:marshaling-model-schematic},
    \ref{fig:marshaling-model-detailed}, a) We first parse the natural language inputs ($\tau = B, O, Q$) into a set of candidate probabilistic program condition and query expressions ($\Pi_O, \Pi_Q$) to be passed on to future model synthesis stages. 

    Specifically, we use an LM that has been prompted to parse each sentence of input natural language observations into a corresponding formal expression ($\pi_O$) intended to condition a probabilistic model with observed constraints on latent variables, and questions into query expressions ($\pi_Q$) that define target variables for inference in that model.This stage draws on prior cognitive modeling work that suggests people use learned statistical distributions to understand language by translating from sentences into symbolic mental modeling representations~\citep{wong2023word}. Unlike in this prior work, this parse stage occurs \textit{before} we have a fully synthesized mental model. To that end, this stage parses natural language into \textit{placeholder} expressions containing functions and variables that will only come to be defined in later stages of model synthesis.

    We sample proposed parses from an LM conditioned on the input task, and prompted with example parses derived from other tasks and corresponding probabilistic models (see Appendix for prompting details):
\begin{equation}
  		 \Pi_O, \Pi_Q \sim P_\text{LM}(\Pi_O, \Pi_Q| \tau)
\end{equation} 
We then score each sampled parse according to an evaluation function $\Phi_{parse}$ (also defined using an LM prompted with example parses). We generate $k_{parse}$ candidates and greedily select the best conditioned on the input task: 
\begin{equation}
  		 \Pi_O^{*}, \Pi_Q^{*} = \text{argmax}_{i \in k_{parse}}\Phi_{parse}(\Pi_O^i, \Pi_Q^i | \tau)
\end{equation} 
This best parse (we use the * notation throughout to refer to the best scoring generations, from a set of candidates, with respect to a utility function $\Phi$) is then passed on to the next stage. 

\item \textbf{Relevant Natural Language Background Description:} (\autoref{fig:marshaling-model-schematic},
    \ref{fig:marshaling-model-detailed}, b)  Next, we retrieve candidates for additional, relevant background knowledge details ($B^+ = \{b^+_1, b^+_2, ... b^+_{N'}\}$). These will be combined with the initial (potentially underspecified) input background $B$ to yield an augmented natural language description 
    $B_{aug} = B \bigcup B^+$ that is intended to fully specify, in explicit detail, latent relevant variables for reasoning about the task at hand. 

    We use an LM that has been prompted to generate detailed background descriptions that name relevant variables given an input task, and explicitly describe their functional form and relationship to other  named variables (e.g., this generation might include sentences describing the prior distribution of a variable and its dependencies on others, like \textit{an athlete's rowing speed varies widely among athletes, but stronger athletes tend to row faster}) (see Appendix for prompting details). This generation is conditioned on both the task input and best parse:
\begin{equation}
  		 B_{aug} \sim P_\text{LM}(b^+_1, b^+_2, ... b^+_{N'} |\Pi_O^{*}, \Pi_Q^{*}, \tau).
\end{equation}

    \item \textbf{Conceptual Dependency Graph:} (\autoref{fig:marshaling-model-schematic},
    \ref{fig:marshaling-model-detailed}, c). Jointly with generating $B_{aug}$, we also generate a conceptual dependency graph G which summarizes the  dependency structure between all variables in $B_{aug}$.

    That is, we prompt the LM to ``use the description and scenario to extract out a list of concepts that your description implies [and] include any dependencies between the concepts.'' We include this stage because we find empirically that program synthesis in our pipeline more accurately reflected the $B_{aug}$ with it.  (We expect this to be less necessary with more sophisticated synthesis approaches or LMs fine-tuned specifically for probabilistic program generation.) After generation, we jointly score the augmented Natural Language Background descriptions $B_{aug}$ and corresponding Conceptual Dependency Graphs using an evaluation function $\Phi_{relevance}$ (defined using an LM prompted with example retrieved variables and graphs).W We generate $k_{relevance}$ candidates and greedily select the best conditioned on the input and all previous stages of generations:
\begin{multline}
      B_{aug}^{*}, G^{*} = \\ 
         \text{argmax}_{i \in k_{relevance}}\Phi_{relevance}(\{B_{aug}, G\}^i | \Pi_O^{*}, \Pi_Q^{*}, \tau)
\end{multline}

    \item \textbf{Probabilistic Model:} (\autoref{fig:marshaling-model-schematic},
    \ref{fig:marshaling-model-detailed}, d) We generate a full symbolic probabilistic model $M_\text{ad-hoc}$.
    Specifically, we prompt the LM to generate a set of background, stochastic probabilistic program function definitions $\Pi_B$ in our probabilistic programming language WebPPL that encode a prior over all of the variables in $B_{aug}^{*}$, based on the description of their functional form and dependencies in language. The generated model should also be capable of be conditioned and queried using the previously generated expressions $\Pi_O*, \Pi_Q*$ parsed from the input observations and questions. That is, the full model should define any placeholder functions that had previously appeared in the $\Pi_O*, \Pi_Q*$, so that collectively the definitions and parses can be combined to yield a valid probabilistic program $M_\text{ad-hoc} = (\Pi_{B}, \Pi_O*, \Pi_Q*)$. We sample candidate models using a prompted LM conditioned on the preceding steps (see Appendix for prompting details):
\begin{equation}
  		M_\text{ad-hoc} \sim P_\text{LM}(M_\text{ad-hoc} | B_{aug}^{*}, G^{*}, \Pi_O^{*}, \Pi_Q^{*}, \tau).
\end{equation}

We evaluate the sampled model for formal validity $\Phi_{\text{model}}$ and finally return the best model as:
\begin{equation}
  		 M_\text{ad-hoc}* = \text{argmax}_{i \in k_{program}}\Phi_{\text{model}}(M_\text{ad-hoc}^i).
\end{equation}
Our $\Phi_{\text{model}}$ is a boolean function for executability (we just return the first synthesized model that compiles and can return inference results under a given inference algorithm), though it could be implemented as a continuous function based on inference cost.

 \item \textbf{Probabilistic Inference:} (\autoref{fig:marshaling-model-schematic},
    \ref{fig:marshaling-model-detailed}, e) Finally, we compute probabilistic inferences in $M_\text{ad-hoc}$ using general Bayesian inference algorithms defined over the probabilistic programming language -- here, we use rejection sampling, though any other algorithms could be used (e.g., MCMC or even separately defined inference programs as in \citep{cusumano2019gen}). We return inferences in the queried and conditioned model. 
		\begin{equation}
  			P(A | M_\text{ad-hoc}*).
    \end{equation}
as a joint distribution over the set of answers $A$ corresponding to input questions.
    \end{itemize}
The supplement (\textit{Model Synthesis Architectures: Additional Implementational Details}) includes a link to the repository with prompting details for the LM models, additional information on each generation stage, and example generated parses, natural language background knowledge, conceptual dependency graphs, and excerpted models.

\section{Natural Language Reasoning Experiments}
\begin{figure*}[t!]
    \centering
    \includegraphics[width=\linewidth]{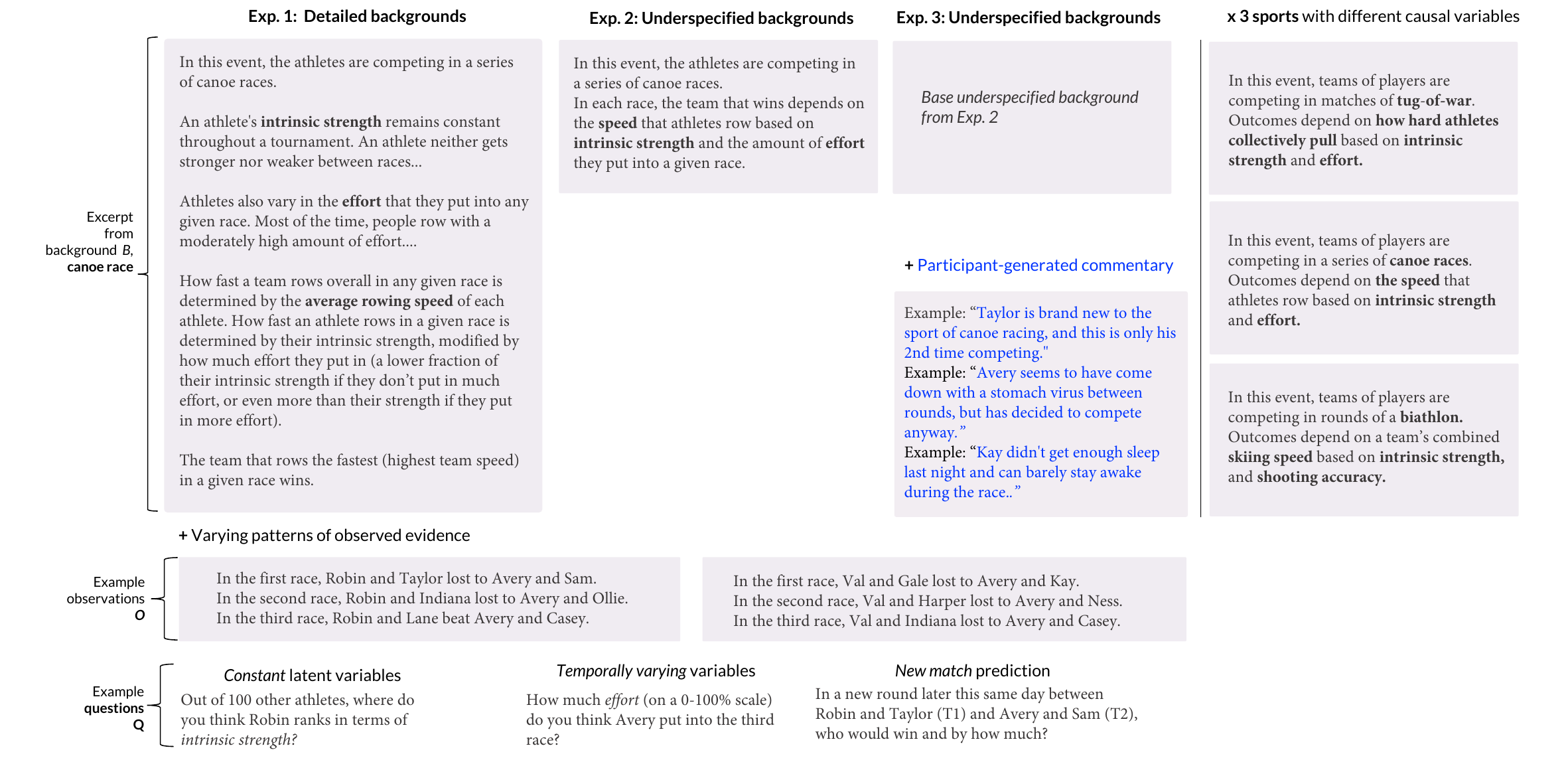}
    \vspace{-0.7cm}
    \caption{Experiment overview for three natural language reasoning experiments. Experiment 1 provides detailed backgrounds spelling out relevant causal variables, their functional forms, and their dependencies; Experiment 2 uses underspecified natural language backgrounds that name relevant variables but not their form or dependency structure; Experiment 3 uses underspecified backgrounds and introduces additional participant-generated ``commentary". Vignettes combine these backgrounds with varying patterns of observed evidence about sports outcomes. Vignettes were presented alongside questions asking about constant latent variables; temporally varying variables; and predictions for new matches involving the same athletes. Vignettes across the experiments spanned three different ``sports" (tug-of-war, canoe racing, and biathlon) with different underlying causal structures.}
    \label{fig:marshaling-experiments}
\end{figure*}

To evaluate flexible ad-hoc reasoning in people and models, we construct a domain of \textit{natural language inference problems}. We then design three experiments around this domain that require people and models to bring to bear progressively more background information to reason about the problem at hand. \textbf{Experiment 1} evaluates ad-hoc reasoning in the most constrained setting, in which essentially all relevant causal relationships are specified explicitly in language. \textbf{Experiment 2} presents vaguer and more underspecified situations, in which key relationships between relevant variables must be retrieved from prior knowledge. \textbf{Experiment 3} introduces new variables, proposed by other naive human subjects, where the causal relationships to existing variables must be constructed by the reasoner, again drawing on background knowledge. In each experiment, a set of 8 questions probes the coherence with which people infer latent variables and predict future outcomes.\\

Here we describe the domain and human experimental methods in more detail, followed by our computational methods for evaluating the MSA implementation and model baselines.

\subsection{Domain -- Model Olympics vignettes} 
Imagine that a friend invites you to their university, where there is an annual, informal boat racing tournament every summer. They ask you to participate, and fill you in on some of the other friends they know will be racing (``Alice used to row crew in high school”, ``Billy and Alice beat Carmen and David in a race earlier today”, ``Elaine and Alice beat Carmen and David yesterday – even though Elaine had one of the older oars.”) You then need to decide who you want to recruit for your team – who is more likely to win? As you hear these observations, you may be continually drawing inferences about the strength or competitive temperament of the players involved (e.g., you may think things like: ``wow, if Elaine won even with the older oars, maybe they try pretty hard'', or ``maybe Carmen and David are especially weak, given their two losses...'') 

While this setting is simple, it nonetheless draws on exactly the kind of everyday, ad-hoc reasoning at the heart of this paper – we naturally come to these probabilistic inferences on-the-fly from partial information about new situations. Most real-world settings that involve reasoning from language are intrinsically ``open-world”, with no natural limit on the kinds of variables and dependencies that could be relevant. 

Our experimental domain is a set of natural language vignettes like these about athletes playing sports. The domain is inspired by the ``Bayesian tug-of-war'' stimuli used to evaluate probabilistic human reasoning in prior work \citep{goodman2014concepts}. In keeping with the focus of our paper, we expand on these problems to test the coherence of people's reasoning as progressively more background information must be integrated to form judgments.

To evaluate how people reason under varying contexts, we design three different background subdomains, each with a different causal structure, underlying our vignettes. These subdomains are three different sporting events: \textit{tug-of-war, canoe racing}, and a \textit{biathlon}. The tug-of-war subdomain largely replicates the one in prior work, involving two latent variables (constant athlete \textit{strength} and temporally-dependent athlete \textit{effort} in each match) that together influence observed outcomes (like who wins particular tug-of-war matches). The canoe-racing  subdomain tests generalization to a novel setting that could not have appeared in LLM training data (unlike the widely published tug-of-war problems), but that otherwise parallels the tug-of-war causal model (again with the constant athlete \textit{strength} and temporally-dependent athlete \textit{effort} that jointly contribute to observed outcomes in canoe races). The biathlon  subdomain is more distinct (involving constant athlete \textit{strength} and temporally-dependent \textit{shooting accuracy} in a combined skiing and shooting event) and intended to be more novel to participants overall. All three sports share a general structure (one constant and one temporally varying latent variable that contribute jointly to performance), allowing us to construct parallel vignettes and inference questions applicable to all three sports. We also formalize the causal structure of each sport into a gold, \textit{hand-crafted probabilistic model} codifying the causal dependencies between these latent variables and observed outcomes between teams. These hand-crafted models provide a computational baseline for reasoning about each sport. 

To evaluate generalization within these contexts, we construct a set of 16 base tournament motifs. Each motif varies the competing teams and observed outcomes in a given tournament to provide different kinds of evidence about the participating athletes (eg. a \textit{round robin} tournament in which players rotate across different teams; a \textit{confounded teammates} setting in which two players consistently play on the same winning or losing team; and several \textit{Bayesian ``explaining away”} settings in which one or two players generally win or lose, but participate in one fluke loss or win, suggesting a competing factor, like temporary lack of effort, for the fluke outcome.) The sixteen motifs draw on the twelve team tournament evidence patterns in the original tug-of-war experiments, which we then augment with four more templates involving noisy, anomalous outcomes to probe probabilistic reasoning.

We generate our experimental dataset of vignettes by independently sampling 6 of these tournament motifs for each sport, resulting in 6x3 = 18 total vignettes. Our generation procedure randomly assigns gender-neutral names to the athletes in each vignette to avoid
any name-based priors about athlete strength. The generation procedure can be used to dynamically generate new vignettes and can be extended with new sports and motifs.

\subsection{Eliciting  probabilistic judgments to probe local coherence}

To probe for local coherence, we want to elicit judgments from people that probe how they reason about \textit{related variables} in a given context. Our core hypothesis predicts that these inferences are consistent with an internally structured, context-specific model over relevant variables. We expect participants' inferences about which athletes are stronger or weaker, or more or less effortful, to coherently relate to each other and to their predictions about future matches. Similarly, if they decide that other variables (injuries, energy drinks, mistakes) are relevant to an observed outcome, then we expect to see a coherent influence of these novel variables on other judgments.

We design a new elicitation interface to address two goals: (1) to elicit probabilistic judgments that reflect structured uncertainty, and (2) to elicit these judgments over related variables and predictions to probe for local coherence. 

To allow participants to express structured uncertainty, we design a \textbf{multi-sample judgment} interface where participants entered $k_{click} = 5$ judgments for each question, which we interpreted as a set of samples $S_{q_i} = \{\hat{a_1},\hat{a_2},...,\hat{a_5}\}$ for each query $q_i$. This interface was inspired by ~\cite{gerstenberg2018happened} and allows participants to express modes in their marginal posterior conditioned on the background and evidence. 

We also designed each of the 16 tournament motifs with a corresponding \textbf{palette of 8 inference questions}, chosen to probe for consistency and coherence across related inferences. These always consisted of three questions about the constant latent variable for three athletes, three questions about the temporally varying latents for the same athletes in specific matches, and two prediction questions about new matches. We design the questions that should reveal coherent and related inferences, based on the tournament motifs and hypothesized causal structure of each sport – for instance, we ask questions about the two teammates who always play together in the \textit{confounded teammates} tournaments, or questions about the anomalous win or loss matches in the \textit{Bayesian ``explaining away”} settings where players experience surprising upsets. When we generate individual vignettes for each sport, we also generate the corresponding palette inference questions, instantiated with the specific constant and temporally relevant latent variables (e.g., \textit{``On a percentage scale, where 0\% is no effort and 100\% is maximum effort, how much effort do you think Val put into the third match?''} for a tug-of-war or canoe-racing vignette, or \textit{``On a percentage scale, where 0\% is zero accuracy and 100\% is perfect accuracy, how accurate do you think Val was at shooting in the third round?'' for a biathlon}). We ask this palette of questions about related variables (as opposed to the single inference question per situation in \citealp{goodman2014concepts}) to probe whether people deliver judgments consisting with a holistic underlying causal representation of each situation.

\subsection{Human experiments}

We next describe the sequence of three experiments using our Model Olympics domain (see \autoref{fig:marshaling-experiments}). Additional details on all experiments, including examples of the vignettes and images of the experimental interface, can be found in the Supplement. 

\begin{itemize}
\item \textbf{Experiment 1 (detailed background context)} tests ad-hoc reasoning about arbitrary collections of variables when they are described explicitly for a given situation. Vignettes are presented with detailed linguistic background descriptions (\autoref{fig:marshaling-experiments}, left) based on the gold, hand-crafted probabilistic models for each sport. These background descriptions spell out the functional form of relevant variable distributions (eg. \textit{An athlete's intrinsic strength remains constant throughout a tournament, and varies widely over the athletes in the tournament, with mostly average strength athletes and a few much weaker and much stronger athletes}) and causal relationships between these variables (eg. how \textit{strength} and \textit{effort} determine an athletes' per-match rowing speed in canoe racing) that ultimately lead to the observed match outcomes. $N_{E1}=78$ participants from Prolific judged a random sample of two vignettes from each of the sports. 

\item \textbf{Experiment 2 (under-specified background context)} tests ad-hoc reasoning when some relevant variables are briefly described or implied in language, but most of the relevant intermediate details must be filled in from a participants’ background knowledge. Vignettes are presented with brief and under-specified background descriptions (\autoref{fig:marshaling-experiments}, center) that name the constant and temporal latent variables in the hand-crafted model for each sport, but do not explicitly spell out the functional form of their underlying distributions or how these variables interact to produce observed outcomes. This setting has not typically been considered in prior controlled computational causal inference models like \cite{goodman2014concepts}, but is a particular focus given our interest in people's ability to retrieve globally relevant information to reason about.  $N_{E2}=80$ participants from Prolific judged the same batches of vignettes as in Experiment 1, allowing comparison across experiments to probe whether people drew similar distributions of inferences to the explicitly detailed setting. 

\item \textbf{Experiment 3 (participant-generated novel details)} tests how people flexibly incorporate arbitrary evidence into ad-hoc reasoning, by introducing uncontrolled new variables from outside the scope of our original domain. To do this, we extend the under-specified vignettes from Experiment 2 with new participant-generated details (“sports commentary”, \autoref{fig:marshaling-experiments}, right), from naive participants prompted to come up with new, relevant observations that would have changed their own predictions about a particular future match (eg. to make a given outcome more or less likely).\\\\ These commentary introduced novel variables (injuries, performance enhancing drugs) that test participants ability to reason out of distribution by extending base stimuli (similar to the experimental paradigm in ~\cite{collins2022structured}). Participant-generated details were collected from $N_{E3,a}=20$ naive human subjects in a separate elicitation task. This experiment used a smaller set of 9 total vignettes across two sports (tug-of-war and canoe racing), with slightly different patterns of evidence than those in Experiment 1 and 2, due to our vignette sampling procedure. In the main judgment task with the extended vignettes, $N_{E3,b}=20$ participants judged all 9 vignettes. 
\end{itemize}

\subsection{Computational experiments}
Across all three experiments, we elicit simulated judgments from the MSA and alternatives in the form of estimated posteriors for the palette of questions, conditioned on the vignettes.

\paragraph{MSA model configuration} 
We instantiate our MSA architecture using \texttt{Llama-3.1-70B} as our base LM for all parsing, code synthesis, and LM-based evaluations; and WebPPL \citep{webppl} as our probabilistic programming language. 

We simulate a ``participant'' under the assumption that they construct a single, situation-specific ad-hoc probabilistic program representing the relevant variables for a given vignette. This instantiates the hypothesis that people sample a single mental model or ``conception'' of the situation at hand when reasoning. Future work might explore whether individuals are better understood as integrating multiple distinct probabilistic models to reason about any one scenario.

At all stages of the pipeline that involve an LM proposal, we prompt the LM with both the specific vignette and randomly sampled background examples showing other vignettes (and corresponding parses, dependency graphs, and probabilistic programs for those vignettes, at each stage of the pipeline). For each vignette, and for each simulated `participant', these examples consist of sample vignettes and models for the other two sports in our domain (e.g., to synthesize a tug-of-war model, the LM is shown examples from the canoe and biathlon sports), and two more examples demonstrating more general features of WebPPL syntax (a vignette about student performance on an exam, and a vignette about a synchronized diving sport that is not included in our domain). This prompting strategy allows us to evaluate whether our model generalizes better when given closely analogous examples (e.g.,  from canoe to tug-of-war) versus more distant inputs (e.g., to the more novel biathlon structure). Additional experimental details including LM prompts can be found in the Supplement.

For each simulated participant and vignette, we estimate posteriors for each question with $k_{samples}=1000$ samples for Exp. 1/Exp. 2 and $k_{samples}=500$ samples for Exp. 3 (drawn using rejection sampling for an unbiased estimator of any model expressed in the PPL). In our results, we simulate $N=10$ participants per vignette. Our overall sampling budget is constrained (and less in the more complex Exp. 3 setting) as rejection sampling incurs a significant time cost.

\paragraph{Alternative models and baselines}
We compare the MSA judgments to those of several alternative models:
\begin{itemize}
\item \textbf{Gold symbolic models:} For Exp. 1 and Exp. 2, we estimate posteriors using the hand-designed, gold symbolic models constructed for each of the three sports. These allow us to evaluate whether people's judgments varied with respect to those in prior work on natural language inference \citep{goodman2014concepts,goodman2024probabilistic}, given our extended experimental settings. This baseline also allowed us to assess how judgments in our MSA-synthesized probabilistic programs compared to those from hand-crafted models. For direct comparison to the MSA implementation, we again simulate 10 participants for each vignette, estimating posteriors for each simulated participant and vignette with $K=1000$ samples (using rejection sampling). 

\item \textbf{Large language models (direct and CoT):} To evaluate the role of symbolic model synthesis, we implement two LM-only alternatives using the base LM model (Llama-3.1-70B). We evaluate a ``direct'' response setting, where we prompt the LM to directly answer all questions for each vignette via  feedforward generation, and a chain-of-thought (CoT) setting \citep{wei2022chain}. We consider the CoT setting to be an ablation of our MSA approach, as it involves greater computation in language (similar to the pipeline stages in MSA that precede probabilistic program synthesis) but no explicit symbolic model execution. Both LM baselines are prompted directly with the experimental instructions shown to human participants and the vignettes. For both baselines, we simulate a participant using the same multi-response paradigm, in which the LM sequentially generates 5 scalar judgments for each of the 8 questions, conditioned on all of its previous answers to that vignette. We generate from the LMs at temperature of 1.0 to elicit response diversity. Lower temperatures are poor fits to the distribution of human responses because the model responses do not vary substantially. 

\end{itemize}
\begin{figure*}[ht!]
    \centering
    \includegraphics[width=\linewidth]{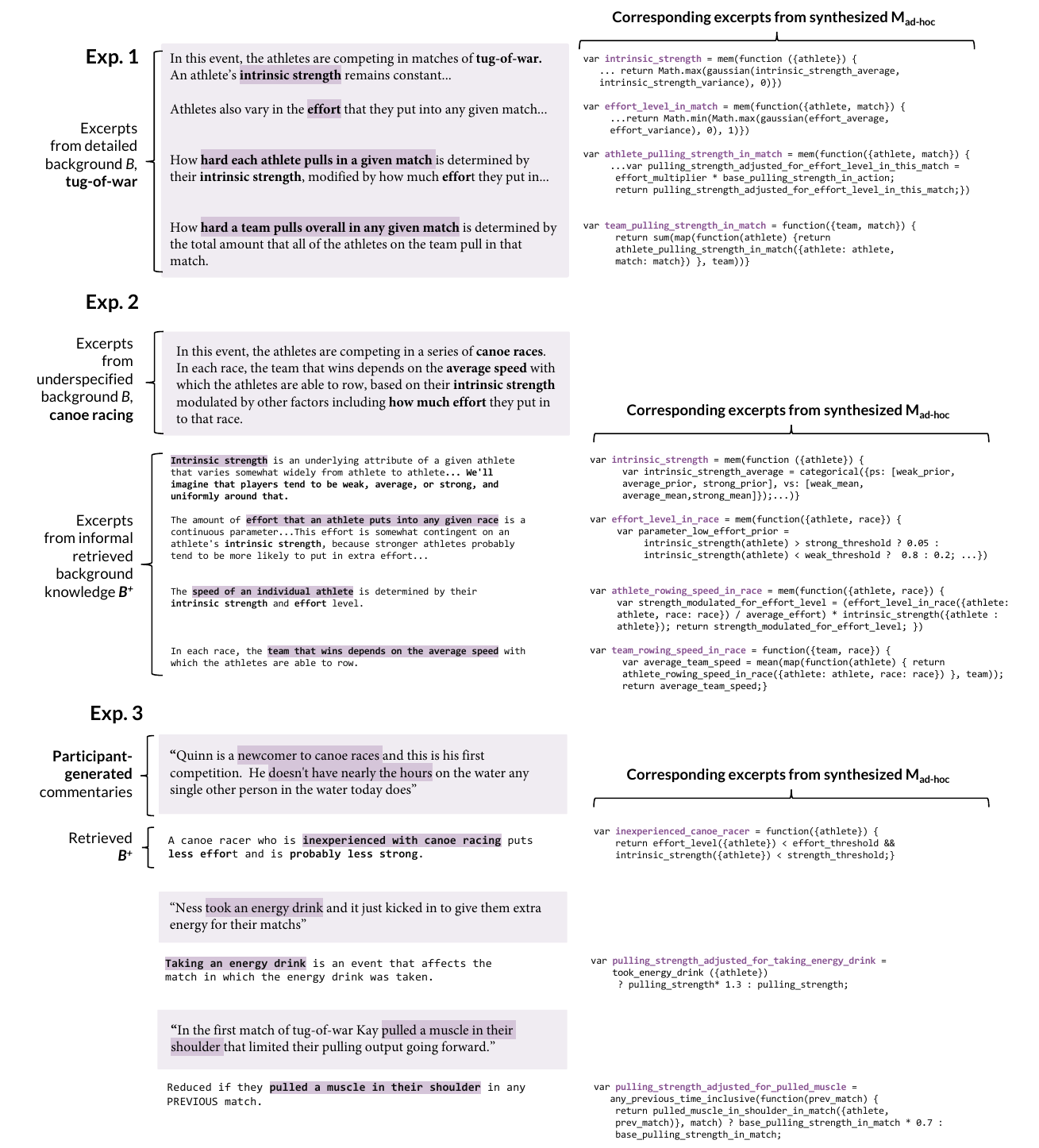}
    \vspace{-0.5cm}
    \caption{Excerpts showing key parts of the \textbf{natural language inputs}, retrieved additional \textbf{informal background knowledge} $B^{+}$ as natural language describing proposed relevant latent variables, and resulting \textbf{formal ad-hoc models} $M_\text{ad-hoc}$ as synthesized probabilistic programs. \textbf{Exp. 1} (top) examples show how explicit natural language descriptions of causal variables (left) are grounded into stochastic function definitions that reflect them (right). \textbf{Exp. 2} (center) shows how our pipeline conditions on \textit{underspecified} backgrounds to retrieve relevant additional causal variables in natural language (left, bottom) that are formalized into resulting synthesized programs (right). \textbf{Exp. 3} (bottom) shows how our pipeline conditions on \textit{additional participant-provided free-form natural language} to also retrieve background causal relations relating them to the observed variables, which are also formalized into ad-hoc model definitions (excerpted on the left).}
    \label{fig:example-synthesized-models}
\end{figure*}

\begin{figure*}[ht!]
    \centering
    \includegraphics[width=\linewidth]{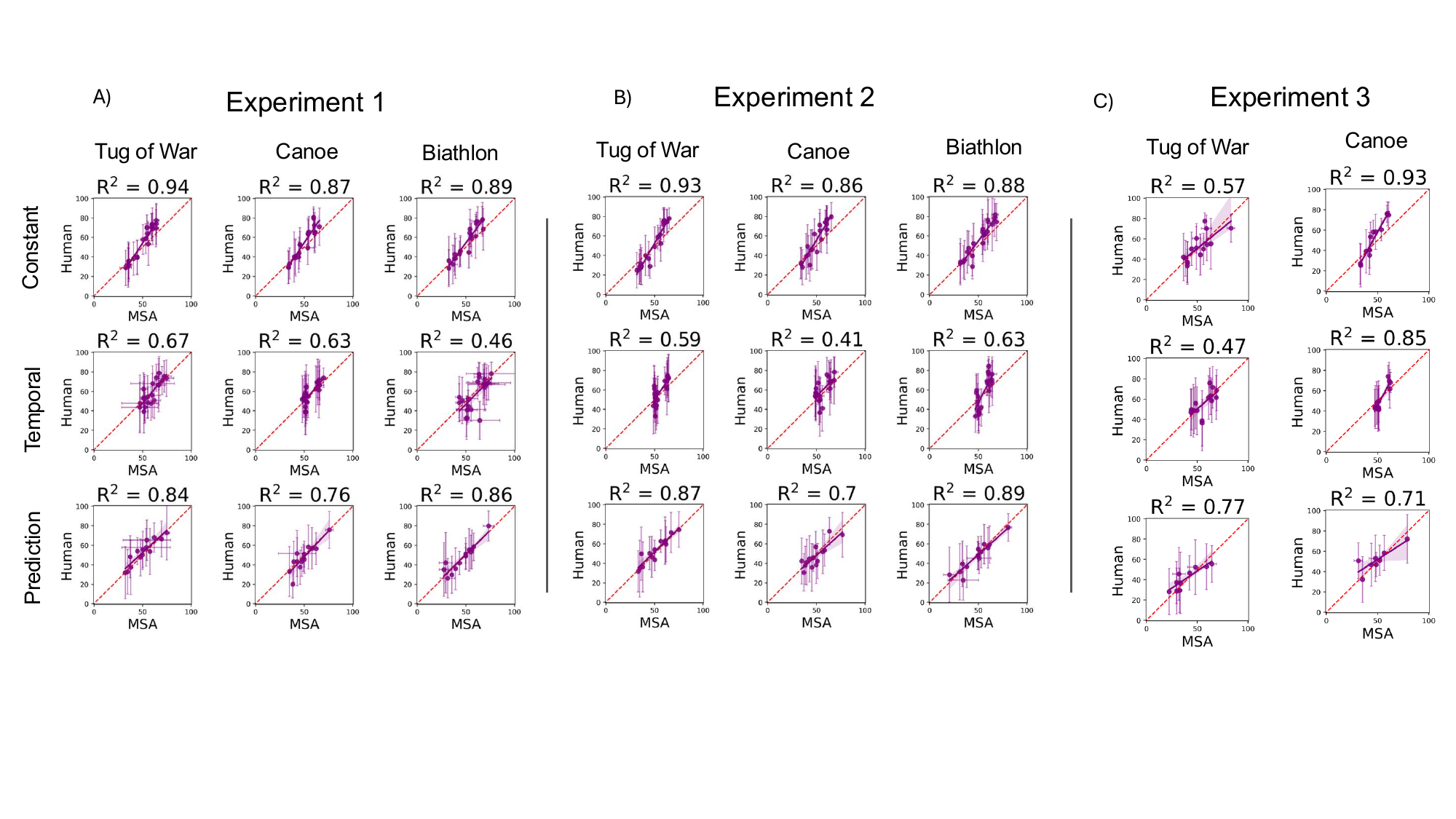}
    \vspace{-2.5cm}
    \caption{Correlations between human judgments and MSA predictions across Experiment 1 \textbf{(A)}, Experiment 2 \textbf{(B)}, and Experiment 3 \textbf{(C)}. Each plot shows correlations within a specific sport and query type (queries about the \textit{constant} latent variable in the top row, queries about the \textit{temporally varying} and match dependent latent variable in middle, and queries about the \textit{new match predictions} in bottom.) Points on each plot are between \textit{mean} predictions for each query, for each scenario, over all participant answers for that query (combining all human clicks into a single posterior before taking the mean) and over all simulated participant model posteriors (combining all model posteriors) for that query.
    Error bars show standard deviation over participants (humans) and simulated participants (models). Additional scatterplots showing human-model judgments for all alternative models and baselines appear in the supplement.}
    \label{fig:scatter-msa-e1-e2-e3}
\end{figure*}

\begin{figure*}[ht!]
    \centering
    \includegraphics[%
    width=0.8\linewidth,
  ]{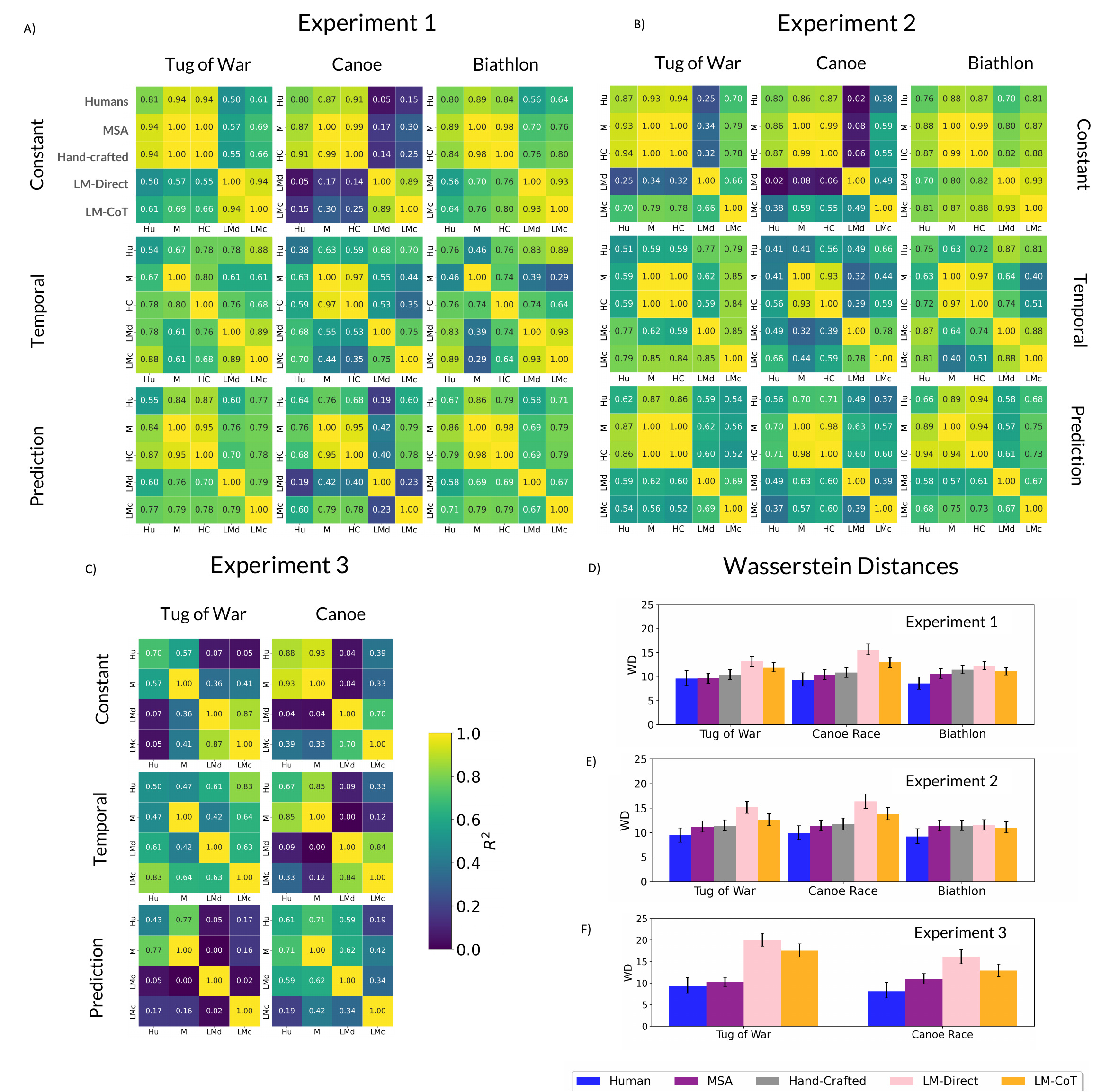}
    \caption{Correlational and distributional comparisons of MSA, hand-crafted symbolic, and LM-only alternative models to human judgments. \textbf{(A)-(C)} Cross-model and model-human $R^2$ for each experiment, where $R^2$ are computed over mean judgments per scenario per query. Fits of the models to people, as well as the structure between models (e.g., the LMs are more similar to each other, in many cases, than to people or MSA). The color bar ranges from $0$ to $1.0$ (R$^2$). Reasoners are abbreviated (Hu=human, M=MSA, HC=hand-crafted symbolic probabilistic program, LMd=LM-Direct, LMc=LM-CoT); \textbf{(D)-(F)} Wasserstein Distances from models and baselines to distribution of human judgments (individual Wasserstein Distances (WDs) computed between judgments per query per scenario, then aggregated as the mean over query types, and mean across query types for each depicted sport and experiment; \textit{lower} distances indicate closer similarity to the distribution of human judgments. Error bars for model-humans show 95\% CI over $1000$ bootstrapped samples, with replacement, on the human data; for human-humans, over $1000$ sampled 50-50 split-half WDs.}
    \label{fig:crosscomp-r2-wasserstein-msa-e1-e2-msa-LM-cot-e3}
\end{figure*}

\subsection{Comparing Probabilistic Judgments} 

We compare human judgments to all models using two distinct measures:

\begin{itemize}
    \item \textbf{Correlational analyses} ($R^2$) between people and models, computed between the mean judgments across all participants (combining all participant clicks for each query) and the mean judgments for each model (combining all simulated participant samples for each query). Correlating mean judgments provides a readily interpretable, aggregate picture of how well models capture human judgments.
    \item \textbf{Distributional measures} (\textit{Wasserstein distances}) offer a more granular picture of fit by comparing the full marginal distributions of human judgments (on the multi-click paradigm) to the full distribution of and model samples~\citep{ying2025benchmarking}. Specifically, we consider the Wasserstein Distance (WD) metric, also known as Earth-Mover's Distance, to compare the similarity in probability distributions over inferred variables -- \textbf{lower} Wasserstein Distances between model judgments and human judgments mean that the distributions are \textit{more similar} to one another. To construct an ``approximate human posterior'' from the human judgments, we treat each person's click (of their 5 clicks) as a sample that lives between $0$ and $100$ (the scale of each query).\footnote{That is, if there are 20 people each clicking 5 times, we have 100 samples that form our ``human posterior.'' We repeat this procedure for the LM baselines.} We then discretize the distribution by bucketizing the queries in $10$ buckets. We then compute the WD between these discretized distributions for each question for each vignette. We also compute the aggregate WD by averaging over the individual WDs. (We also consider an alternate distributional measure, Total Variation Distance, in the Supplement.) 
\end{itemize}
We also compute \textbf{split-half comparisons between human participants} (randomly splitting the participants into equal-sized batches and computing correlations and distributional distances between the participants in each batch, with 95\% confidence intervals bootstrapped over 1000 random samples of these batches). All of our analyses combine judgments across participants, so the split-half human-human comparisons provide a noise ceiling on the explainable variance across the sampled participant population, and a baseline for judging the quality of model fits on each metric.

\section{Results}
Using both these correlational and distributional analyses, we assess how people and models reason in each of our three experimental settings. We collate our results into a series of key findings. 

\paragraph{Key finding 1: People’s reasoning is generally consistent with Bayesian inference in ad-hoc probabilistic models.}
\autoref{fig:scatter-msa-e1-e2-e3} and \autoref{fig:crosscomp-r2-wasserstein-msa-e1-e2-msa-LM-cot-e3} (A-C) shows that across all three experiments, inferences in the probabilistic models synthesized by our MSA are generally well-correlated with human judgments. The supplemental results include full scatterplots like those in \autoref{fig:scatter-msa-e1-e2-e3} for all correlational analyses, and further show that these fits generally fall within the 95\% CI range of the split-half human-human baseline, when bootstrapped over random batches of participants. 

This result is also borne out by our distributional analyses. \autoref{fig:crosscomp-r2-wasserstein-msa-e1-e2-msa-LM-cot-e3} (D-F) shows that the MSA captures not just the average human judgment, but often the distribution of predictions. Under this metric, comparison to the split-half, human-human baseline again shows that the distribution of MSA inferences is about as similar to human judgments (\textit{purple}) as human participants are to each other (\textit{blue}). To ensure that this result is robust to different measures of distributional similarity and not an artifact of the Wasserstein Distance metric, we also repeat this analysis with another metric for comparing distributions (Total Variation Distance, shown in the supplemental) and find the same results relative to the human-human baseline. 

Together these results suggest that in each of these experimental settings people reason in ways that are consistent with normative Bayesian inferences over some structured set of relevant variables – and that these variables can be retrieved automatically using our approach. This overall finding has a number of implications, relative to prior work. The results of Experiment 1 essentially replicate prior work \citep{goodman2014concepts} – but show that the relevant probabilistic models can be synthesized directly from the natural language inputs, rather than designed by hand. In this way, the full MSA represents a complete picture of what could be happening in participants' minds, from natural language input to final outputs. The results of Experiment 2 demonstrate the generalizability of this approach to a more naturalistic setting– even when people are not told explicit, structured dependencies between the variables, they still reason in ways consistent with synthesized probabilistic models. Results from Experiment 3 validate the generalizability of this finding to the open-ended setting, where models and participants must reason about novel variables and causal dependencies, beyond those representable in the base model. This finding is particularly important as a partial validation of the scalability of the mental models hypothesis – which requires being able to square mental models with the open-world reasoning people often exhibit. Taken together, these suggest that ad-hoc probabilistic models defined over a limited subset of relevant variables can explain people's judgments as well as previous, fixed model approaches, and that these models can be empirically recovered from natural language inputs. 

\paragraph{Key finding 2: People's reasoning is more similar to inference in structured probabilistic models than LM-only alternatives, especially when generalizing to arbitrary new details.}
We next compare people's inferences to alternate models and compare models to one another to probe the structural cross-similarity in probabilistic judgments across model classes. The blocks of correlations visible on the heatmaps in \autoref{fig:crosscomp-r2-wasserstein-msa-e1-e2-msa-LM-cot-e3} (A-C) show that in many cases, humans are better correlated with themselves and symbolic models (MSAs and the hand-crafted probabilistic models) than with LMs, which are instead better correlated with each other. This finding parallels cross-model results from the distributional metric \autoref{fig:crosscomp-r2-wasserstein-msa-e1-e2-msa-LM-cot-e3} (D-F), which also shows that humans, MSA, and hand-crafted models distributions (blue, purple, and silver) are generally closer to the human distribution than the LM distribution (pink and orange). 

Digging deeper, we find that the differences between blocks of correlations (the people and symbolic models, relative to the LMs) are more pronounced in the canoe race subdomain than the tug-of-war subdomain (compare \autoref{fig:crosscomp-r2-wasserstein-msa-e1-e2-msa-LM-cot-e3}A and B, \textit{Tug of War} and \textit{Canoe} columns). This is notable because the canoe race subdomain was designed to have a highly similar causal structure to the tug-of-war subdomain, which should suggest similar behavior to that subdomain. One difference is that the canoe race subdomain could not have been retrieved from any previously published work that may have been in LM training data. Pure LM models may have struggled in this less familiar setting, while the use of explicit probabilistic models may have improved generalization performance. (In contrast, in the biathlon subdomain, symbolic models and LMs performed comparably in their fit to human judgments. This finding warrants further analysis, but one possibility is that this was an effect of unfamiliarity with the subdomain on the part of participants – who consistently reported that they were not familiar with the sport. Both LMs and MSAs, via their LM components, may have been more familiar with the subdomain -- understanding the dependencies that exist between its variables -- and accordingly have behaved more similarly to one another.)

In Experiment 3, which targeted the open-world setting, MSA judgments were substantially better aligned with people's than were those of LM-only alternatives.  That these differences were greatest in this setting is suggestive in two ways. First, LM baselines may face particular challenges in fitting human judgments as the distribution shifts further from familiar settings. This interpretation parallels findings from ~\cite{collins2022structured}, which found that LLMs deviated from people more when reasoning about progressively more novel scenarios.

Second, as noted above, this experimental setting represents a significant a priori challenge to existing, hand-crafted symbolic models of cognition -- which generally cannot operate in an open-world setting where new variables and dependencies become relevant as new observations are introduced. Indeed, we do not have a hand-crafted symbolic model baseline in Experiment 3 for precisely this reason -- the original hand-crafted models would not have been able to account for the new observations and relevant variables introduced by human participants. That the MSA continues to outperform LM baselines in this case that probabilistic modeling can continue to best capture human judgments even in the open-world setting.

Globally, these findings suggest an asymmetry in LMs abilities – LM's may be relatively better equipped to retrieve relevant world knowledge and causal dependencies in human-like ways, but relatively less able to integrate that evidence into a locally coherent world model the way that people do. 

\paragraph{Key finding 3: MSA can retrieve and represent relevant information about arbitrary situations as structured probabilistic models.}

Both correlational and distributional analyses suggest that the implemented MSA can synthesize models that quantitatively capture human inferences. But what do these ad-hoc probabilistic models actually look like – what relevant background knowledge do they surface, and how do they represent the content of linguistic background knowledge (both in the original inputs, and retrieved through our approach) as probabilistic programs? The qualitative examples in \autoref{fig:example-synthesized-models} show that the LM-guided synthesis approach is generally able to retrieve reasonable descriptions of variables and causal dependencies (e.g, that intrinsic strength might break down along several modes amongst athletes, or that per-match effort is a continuous parameter that might correlate with an athlete’s original strength); and that it can then parse natural language descriptions from both the original inputs (\autoref{fig:example-synthesized-models}, Exp. 1) and retrieved background knowledge (\autoref{fig:example-synthesized-models}, Exp. 2, 3) into corresponding probabilistic programs that capture these variables and dependencies as interrelated function primitives. 

At the same time, manual inspection of the underlying code also reveals places where the MSA implementation generates imperfect parses from natural language to probabilistic programs. These range from somewhat minor (e.g., the retrieved additional natural language $B^{+}$ suggests that the winning team depends on the \textit{average} speed with which the athletes row, but the synthesized model sometimes does not encode this with a \texttt{mean}); to relatively more drastic omissions (for instance, the synthesized models in Exp. 3 often did not correctly interpret modal temporal logics, like that a pulled shoulder limits pulling strength in \textit{future matches}, until we allowed the synthesis procedure to access a library of modal logic functions). We discuss both these limitations and opportunities for further work below. 

\section{Discussion}

In this work, we investigated how people are able to reason in ways that deliver \textit{global relevance} and \textit{local coherence} -- that is, how human reasoning is able to show both a sensitivity to relevant considerations from across people's background knowledge and coherent integration of evidence over those considerations. In our experiments, we found that an MSA can synthesize ad-hoc models that fit human judgments in the first instance, and fit those judgments better than LM baselines. This suggests that MSAs offer a promising avenue towards capturing the computations underlying human reasoning, especially in open-world settings.

\subsection{Reasons For Model Fit}

Why might human judgments better align with MSAs than with LM baselines in these cases? One possibility is that the difference is due to the way both model classes handle coherence. The mental models generated by MSAs are coherent by design, while LMs internal representations do not have similar coherence constraints. If people's judgments over multiple variables tend to be more internally coherent, this could drive the fit to MSAs over LMs. Another possibility is that MSAs use of explicit causal and probabilistic representations might force the model to place more weight on deeper structural properties, rather than superficial features of the language used to describe tasks. If people's judgments are tracking these deeper causal properties of the stimuli, this could explain the better match to MSAs. Such an explanation would fit with similar findings that point to a lack of robustness in these models in response to surface-level features \citep{mccoy2023embers, valmeekam2023planbench, mirzadeh2024gsm}. Determining which of these or other explanations is most plausible, and if this general trend continues to hold in more varied domains, is a priority for future research. 

\subsection{Handling Surprising Evidence}

In our data, people appeared to be close to rational in their integration of evidence with background beliefs, as measured by fit to our MSA. This included integrating unexpected observations (e.g., a surprising  win by a suspected slow runner against a suspected fast runner) in a measured fashion. In cases where LMs differ most from people, a tentative analysis suggests that one of the key challenges faced by LMs was an over-sensitivity to these surprising observations. For example, from qualitative inspection, we noticed instances where the LM baselines tended to believe that a fast runner's single loss to an otherwise slower runner was often enough to neutralize or reverse the model's assessment of their relative speeds, even when the weight of the rest of the evidence suggested otherwise. The tendency of our MSA not to over-index in these cases may be due to the construction of an explicit model, with priors and a causal structure that grounds the integration of competing observations. Further work should explore this theme of holistic integration more thoroughly, including in cases where information is revealed piecemeal over time (as it often is in naturalistic reasoning tasks), rather than all at once (as in our experiments here). 


\subsection{Open-World Reasoning}

Data from Exp. 3 demonstrated the largest differences between model classes in fit to human data. This experiment focused on generalization in the open-world setting, conditioning on participant-sourced commentaries introducing novel considerations. Performance on this experiment represents a particularly interesting kind of generalization -- to observations that require introducing new variables and dependencies into the underlying causal structure, thereby expanding the expressivity of the model (relative to what would have been synthesized in the absence of the commentary; the models synthesized in Exp. 1 and Exp. 2). As noted earlier, reasoning in this open-world setting represents a strong challenge for classical Bayesian models of cognition, which cannot handle novel variables. Despite this, our MSA strongly outperforms LMs in modeling human judgments for these stimuli, suggesting a continued benefit from being able to rely on the kinds of representations that figure in probabilistic models. In particular, MSAs' ability to recombine symbolic representations of the relevant causal structures may have supported a greater degree of generalization to highly novel circumstances. A priority for future work is explore where this ability breaks with LM-powered model synthesis to explore whether other kinds of MSAs might better fit human cognitive abilities in turn. 

\subsection{Distributions of Human Judgments Reveal Structure That All Models Fail To Capture}

One of the advantages of collecting and analyzing distributional data is that we can analyze human and model judgments in more fine-grained ways than conventional measures like $R^2$ allow. A cursory analysis of this data reveals interesting differences between people and both model classes, highlighting the amount of structure in human judgments still to be explained. Compared to human participants, for example, LMs appear to be more streaky – clustering their judgments around particular outcomes – and respond too strongly to surprising observations – yielding judgments that are at times wholly in the opposite direction of people’s. MSAs are more often directionally correct (as evidenced by greater $R^2$), but tend to produce judgments that are visibly smoother and more uncertain than people’s (see \autoref{fig:qualitative-histograms} in Supplement). In short, human judgments appear to have strong opinions (visible streaks, like in the LMs), but place those peaks more consistently over modes in the Bayesian posterior (as evidenced by our MSA’s superior $R^2$ and WD measures).

A pressing question then is whether some other model class could better delivers the patterns seen in the human data. This might be some deeper hybridization of neural and symbolic methods -- one that reproduces the sharply peaked opinions of LMs, but places those peaks more consistently in the right places -- or an MSA with stronger sampling methods that focus samples more directly over modes. Modeling such fine-grained distributional features of human judgments is a target for future work. 

\subsection{Experimental Limitations}

One limitation of the current work is that human data were relatively noisy -- both split half human-human correlations and model-human correlations showed wide confidence intervals. We can also explore ways to make human vairance more model-able – by matching particular mental models (in MSAs) or response patterns (in LMs) to particular participants – to capture individual participant's unique conception of the situtation, for example. 

Variance in samples from our MSA was also often too low. Judgments in Experiments 1 and 2, for example, were highly correlated for our MSA, but not nearly so correlated for people. Similarly, people's judgments in certain conditions, such as in the canoe domain, were often higher variance than those of our MSA. This suggests a lack of diversity in the models synthesized by our MSA. Follow-on work should explore how to increase the diversity of synthesized models, by increasing the number of models, for example, or by more targeted methods, such as conditioning model generation (and LM responses) on participants' self-reports about what they are thinking about. 

Another near-term target for follow-up work is exploring stronger baselines and more thorough model ablations. Anecdotally, we found that a staged model synthesis procedure worked best, but this should be explored systematically and compared to other model synthesis strategies. Similarly, MSA performance should be compared to state-of-the-art reasoning models, as well as the cognitive models derived from them \citep{binz2024centaur}. Leading reasoning models in particular are likely to perform better at these tasks, but also likely to synthesize better probabilistic models if used internally to our MSA. It will be important to see how those two effects wash out when both are compared for human-likeness. 

\subsection{Limitations in Model Synthesis}

Much like the LMs, our implementation of an MSA also faced important limitations in its ability to generalize. Model generations were often overly influenced by the example models given in our prompt, with a consequent lack in model diversity. For example, while our MSA was often able to reconfigure the primitives in the prompted models into models for the novel sport, it struggled to invent new primitives when these were called for. In Exp. 3 our MSA struggled to make sense of temporal information frequently given in commentaries (e.g., ``Kai was fast until he rolled his ankle in match 4'') until we included an example of the relevant abstraction, a temporal ordering of events, in the prompted models. Once armed with this abstraction, the MSA could model the influence of events before, after, or during, but it struggled to build these abstractions on its own. Some of these issues of prompt sensitivity might be ameliorated by using larger LMs or models specially fine-tuned for the task of model synthesis, which might learn to more systematically explore the space of possible models. 

\section{Related Work}

The current work relates most closely to four lines of work, on model approximation and the Frame Problem, on LM guided model synthesis, hybrid models of language comprehension, and LM primitives in probabilistic programs. 

\subsection{Model Approximation \& the Frame Problem}
Previous work on the Frame Problem has made significant progress in defining \textit{resource rational} \citep{lieder2020resource} objectives by which small, task-specific models can be constructed to approximate reasoning and planning with respect to larger models or priors \citep{icard2015resource, ho2022people}. This work provides an important theoretical existence proof, demonstrating that it is possible to construct smaller tractable models that approximate larger (and intractable) ones, and that people empirically \citep{ho2022people} show behaviors consistent with these approximations in reasoning and planning tasks. Both of these works have relatively little to say about \textit{how} minds arrive at these smaller approximations. 
The current approach builds on this work by examining, at a Marr algorithmic level, how the mind might construct these models -- by decomposing the process into a relevance-based synthesis procedure, and by showing that this can be instantiated concretely by exploiting learnable patterns acquired from joint program and language experience. 

\subsection{Hybrid Models of Language Comprehension}
Our concrete computational approach is more closely related to work in cognitive science that shows how language models can be used to synthesize probabilistic programs from language, by translating between natural language and a symbolic LoT \citep{wong2023word, zhang2023grounded, ying2023neuro, ying2025understanding,ying2025languageinformedsynthesisrationalagent}. This prior work considers cases where natural language explicitly spells out all relevant symbolic structure necessary for language interpretation. We build on these approaches by by extending model construction to areas where relevant knowledge must be recruited from large bodies of real-world background information, forcing us to confront the challenges of relevance-based retrieval that open-world reasoning poses. 

\subsection{Language Models for Model Discovery}
Our work connects to three related lines of work using code language models to synthesize structured models of the world or behavior. These lines of work differ in the goals of model synthesis, and the symbolic substrate of models they synthesize. 

One thread focuses on using language models to synthesize explicit, symbolic computational cognitive models of 
human~\citep{rmus2025generatingcomputationalcognitivemodels} or non-human animal behavior~\citep{castro2025discovering}. We differ from these works in our focus on synthesizing \textit{probabilistic programs} as the key representational structure for representing ad-hoc models, which affords a particularly expressive model and automatic reasoning class with strong connections to earlier probabilistic modeling work in computational cognitive science. Our work is also somewhat different in its framing and goals. Both earlier works seek discover symbolic cognitive models to automate the proposal of scientific models for studying behavior. While our approach can be interpreted this way, the MSA architecture also represents an algorithmic hypothesis about \textit{how humans minds actually reason}, framing flexible cognition itself as a process of ad-hoc model synthesis.

More broadly, our focus is on modeling how people reason about arbitrary, open-world situations differentiates -- as a proof of concept towards more domain-general cognitive model synthesis over probabilistic models. This differentiates our work from other recent automated model synthesis methods in both cognitive science and AI that have focused on more domain-specific models, such as synthesizing  models to explain social reasoning~\citep{zhang2025autotom,cross2024hypothetical}. Other recent AI work has focused on synthesizing world models that represent (often deterministic) transition functions for decision making and planning~\citep{wang2023voyager,wong2024learning,tang2024worldcoder,piriyakulkij2025poe}. This work could be productively combined with ours to synthesize probabilistic models that support planning and inference to explain an even wider class of ad-hoc reasoning.

Finally, a related and concurrent line of work in AI has begun to use language models to synthesize probabilistic models \citep{feng2024bird,xia2024let}, including probabilistic programs~\citep{li2024automated, domke2025large}. These works are most similar to ours in their formalism, but differ significantly in their goals. The latter works especially focus on automating scientific modeling for statistical analysis from data.  We focus on an expressive probabilistic programming language class designed for cognitive modeling, and evaluate our approach with respect to empirical evidence of human reasoning. However, as with other work on automated modeling, there are rich synergies between these approaches -- such as extending the MSA approach to capture human scientific discovery, or collaborative scientific discovery between AI and human ``thought partners'' that includes jointly modeling a human scientist along with models of the world ~\citep{collins2024building}. 

\section{Looking Forward}

The problem of open-world cognition is the challenge of being able to reason well-enough in the vast space of problems we encounter. We've taken a small step in that direction by showing that reasonable mental models can be automatically synthesized for new problem instances in a novel family of tasks. Much more is needed to determine whether this approach can scale to the level of generality and flexibility seen in human cognition. 

Answering that will require exploring a broader space of possible MSAs. This might include synthesis using other modeling languages that support long-horizon planning \citep{zhi2022pddl}, multi-agent reasoning \citep{chandra2025theories}, or distributional primitives learned from experience \citep{lew2020leveraging,dohan2022language,grand2025self}. Future work should also explore other model synthesis strategies, such as those that refine initial models with external feedback \citep{wong2024learning,wang2023voyager} or that consider multiple models at once \citep{loula2025syntactic}. Finally, future MSAs should \textit{learn} from model construction over time, by components of the synthesis architecture based on previous successes or failures, and by augmenting the modeling language with successful concepts \citep{ellis2021dreamcoder,grandlilo}.  

Future MSAs can be evaluated both based on ground truth accuracy – whether the models they synthesize are any good – and match to various measures of human behavior. We can ask, for example, whether certain modeling languages better capture the generalizations that people endorse, or which synthesis strategies fit the dynamics of human thought processes, as measured by reaction times or systematic shifts in people's judgments. 

The current era of highly general AI systems means that a deep understanding of how human open-world cognition works may now be within reach. We don't yet have a settled view of how people are able to reason in locally coherent and globally relevant ways about the large and ever-expanding space of things people think about, but the way to investigate this is becoming clear. By scaling MSAs, as well as their pure LM alternatives, and systematically comparing them to human data, we can now begin to meaningfully adjudicate between models of human general cognition. Cognitive science has shed tremendous light on how parts of the mind work. It can now begin to study how those parts fit together. 

\section{Acknowledgments}
We thank Kartik Chandra, Noah Goodman, Thomas Icard, Sydney Levine, Josh Knobe, L.A. Paul, Michael Li, Leshem Choshen,  Vikash Mansinghka, Judy Fan, and Gabe Grand for valuable conversations that informed this work. This work was supported by AFOSR (YIP FA9550-23-1-0127), the ONR Science of AI program (N00014-23-1-2355), a Schmidt AI2050 Fellowship to JBT, and the Siegel Family Quest for Intelligence at MIT. KMC acknowledges support from the Cambridge Trust and King's College Cambridge. LCW acknowledges support from a Stanford HAI Fellowship. JDA acknowledges project support by Intel and the National Science Foundation under grants CCF-2217064 and IIS-2238240. Research was additionally sponsored by the Department of the Air Force Artificial Intelligence Accelerator and was accomplished under Cooperative Agreement Number FA8750-19-2-1000. The views and conclusions contained in this document are those of the authors and should not be interpreted as representing the official policies, either expressed or implied, of the Department of the Air Force or the U.S. Government. The U.S. Government is authorized to reproduce and distribute reprints for Government purposes notwithstanding any copyright notation herein. AW  acknowledges  support  from  a  Turing  AI  Fellowship  under grant  EP/V025279/1, The Alan Turing Institute, and the Leverhulme Trust via CFI. This work is supported (in part) by ELSA - European Lighthouse on Secure and Safe AI funded by the European Union under grant agreement No. 101070617. Views and opinions expressed are however those of the author(s) only and do not necessarily reflect those of the European Union or European Commission. 

\bibliographystyle{apacite}
\setlength{\bibleftmargin}{.125in}
\setlength{\bibindent}{-\bibleftmargin}

\bibliography{references}

\onecolumn
\section{Supplement: Modeling Open-World Cognition as On-Demand Synthesis of Probabilistic Models}
\vspace{15pt}

Experiment and model implementation details reference the repository at: \url{https://github.com/lio-wong/msa-cogsci-2025-data}.\\

\subsection{Model Synthesis Architectures: Additional Implementational Details}

\noindent As described in the main text, we sequentially construct $M_\text{ad-hoc}$ in a staged process that interleaves generation and evaluation steps. The base LM used in all experiments is the HuggingFace \texttt{meta-llama/Meta-Llama-3.1-70B-Instruct-Turbo} release. We query the model using the Together API. Here we provide additional parameters and prompting details for each of these stages.\\

\noindent In our experiments, as we described in the main text, we model each \textit{simulated human participant} as ultimately synthesizing a single model $M_\text{ad-hoc}$ conditioned on an input natural language scenario. The following describes the parameterization used for each single simulated human participant. \\

\noindent Each stage of generation involves a frame prompt for that stage, into which we inject a shuffled set of background examples demonstrating each stage of this pipeline for a set of held-out example scenarios (none of which appear verbatim in our main experiments.) Specifically, we use a \textit{held-out} prompting scheme for selecting these examples, where for a scenario from any given sporting domain (eg. \textit{tug-of-war}) we automatically select background examples only constructed for the \textit{other} sports -- in this case, \textit{canoe-racing} and \textit{biathlon}, along with two other example scenarios, \textit{diving} and \textit{exam}, that we use as examples for all scenarios.) \\

\noindent Below, we describe where in the repository one can find the \textit{frame prompts} for each stage, which include a  \texttt{<SHUFFLED EXAMPLES>} token indicating where these shuffled example generations appear. The full set of shuffled examples themselves can be found at the \texttt{example-scenarios} directory at our data repository, which includes:
\vspace{-5pt}
\begin{itemize}
\itemsep0em 
    \item Base \{\texttt{tug-of-war, canoe-racing, biathlon, diving, exam}\}  examples used for \textbf{Exp. 1} and \textbf{Exp. 2}.
    \item Base \{\texttt{tug-of-war, canoe-racing, biathlon, diving}\} examples for \textbf{Exp. 3}. This experiment was run later and we constructed extended examples demonstrating models with free-form additional natural language observations. We also omit the exam example domain from these experiments. However, future work will explore the effect of these examples on generation and seek to construct a more general set of examples (or fine-tune models so that example-based prompting is not necessary; we use it here as we build on a generic base model.)
\end{itemize}
\vspace{-5pt}
Note that these shuffled example text files contain a concatenated set of \textit{all} of the generation stages (eg. each example file contains an example input scenario, parse, background information in natural language, dependency graph, and full probabilistic program.\\ 

\noindent All frame prompts for each generation stage appear under the \texttt{msa-frame-prompts} directory. Generating of the parsing and background-knowledge/dependency graph used a single system prompt which is included in the same directory. No system prompt was used for the ad-hoc probabilistic program model generation stage.
\paragraph{Parsing}
In our experiments, we forward sample only $k_{parse}=1$ parse at \texttt{temp=0.2}. Throughout, we use lower temperatures for generation stages that require greater syntactic control (like code generation) and higher temperatures for tasks that involve generating natural language (like retrieving and generating informal relevant variables.) We also implement an LLM-based evaluation function $\Phi_{parse}$ which scores parses, but as we only take  $k_{parse}=1$ sample per participant this is of limited utility (we find empirically that parse variability is less important for downstream model quality than diversity in informal knowledge generation, but $k_{parse}$ could be increased for more ambiguous and freeform language in future experiments. \\

\noindent The full frame prompt for the parsing stage can be found at \texttt{generate-parsing} in the frame prompts directory and the evaluation prompt can be found at \texttt{score-parsing}.The frame prompt for this stage was injected with shuffled and concatenated examples starting from the input scenario up to the example parses (delimited by \texttt{<START\_LANGUAGE\_TO\_WEBPPL\_CODE>}).\\

\noindent Here we show a few example parses for canoe-racing and biathlon scenarios in \textbf{Exp. 1} and \textbf{Exp. 2} (as only the background information changed between these experiments, the outcome evidence and questions shown were matched for scenarios in Exp. 1 and Exp. 2). We omit a tug-of-war example as the latent variables as it uses the same outcome and latent variable format as canoe-racing. Parses are excerpted from the full scenario, but show examples of a sentence in natural language parsed into a corresponding line of code. Note that the parse code invariably includes calls to placeholder functions that have not yet been generated and must be generated in the final model.

\begin{tcolorbox}[title=\textbf{Example parse for Exp. 1, canoe-racing}, colback=white,colframe=white!50!black]
\textbf{In the first race, Fey and Ollie lost to Lane and Jamie.}
\begin{lstlisting}
condition(lost({team1: ['fey', 'ollie'], team2: ['lane', 'jamie'], race: 1}))\end{lstlisting}

\textbf{Query 1: Out of 100 random athletes, where do you think Fey ranks in terms of intrinsic strength?}
\begin{lstlisting}
intrinsic_strength_rank({athlete: 'fey', out_of_n_athletes: 100})\end{lstlisting}

\textbf{On a percentage scale from 0 to 100\%, how much effort do you think Fey put into the second race?}
\begin{lstlisting}
effort_level_in_race({athlete: 'fey', race: 2})
\end{lstlisting}

\textbf{In a new race later this same day between Fey and Ollie (Team 1) and Harper and Gale (Team 2), who would win and by how much?}
\begin{lstlisting}
who_would_win_by_how_much({team1: ['fey', 'ollie'], team2: ['harper', 'gale'], race: 4})i
\end{lstlisting}
\end{tcolorbox}

\begin{tcolorbox}[title=\textbf{Excerpted parse examples for Exp. 1, biathlon}, colback=white,colframe=white!50!black]
\textbf{In the first round, Robin and Ollie beat Lane and Ness.}
\begin{lstlisting}
condition(beat({team1: ['robin', 'ollie'], team2: ['lane', 'ness'], round: 1}))
\end{lstlisting}

\textbf{Out of 100 random athletes, where do you think Robin ranks in terms of intrinsic strength?}
\begin{lstlisting}
intrinsic_strength_rank({athlete: 'robin', out_of_n_athletes: 100})\end{lstlisting}

\textbf{On a percentage scale from 0 to 100\%, how accurate do you think Robin was at shooting in the second round?}
\begin{lstlisting}
shooting_accuracy_in_round({athlete: 'robin', round: 2})
\end{lstlisting}

\textbf{ In a new round later this same day between Robin and Ollie (Team 1) and Lane and Taylor (Team 2), who would win and by how much?}
\begin{lstlisting}
who_would_win_by_how_much({team1: ['robin', 'ollie'], team2: ['lane', 'taylor'], round: 4})
\end{lstlisting}
\end{tcolorbox}
\clearpage
\noindent Here we show a few excerpted example parses for \textbf{Exp. 3}, specifically showing parses of the free-form participant-provided observations. As there is more variability across these parses, we show several instances of parses sample for different simulated participants to demonstrate variability.

\begin{tcolorbox}[title=\textbf{Example parses for Exp. 3, participant-generated details}, colback=white,colframe=white!50!black]
\textbf{Taylor is brand new to the sport of canoe racing, and this is only his 2nd time competing.}
\begin{lstlisting}
Sampled parse 1: condition(is_brand_new_to_canoe_racing({athlete: 'taylor'}) && is_second_time_competing({athlete: 'taylor'}))

Sampled parse 2:  condition(is_brand_new_to_canoe_racing({athlete: 'taylor'}) && is_only_second_time_competing({athlete: 'taylor'}))
\end{lstlisting}

\textbf{Kay didn't get enough sleep last night and can barely stay awake during the race.}
\begin{lstlisting}
Sampled parse 1: condition(!got_enough_sleep_last_night({athlete: 'kay'}) && barely_staying_awake_during_race({athlete: 'kay'}))

Sampled parse 2: condition(didnt_get_enough_sleep_last_night({athlete: 'kay'}))
\end{lstlisting}

\textbf{In the first match of tug-of-war Kay, while managing to pull off the win from Avery, pulled a muscle in their shoulder that limited their pulling output going forward.}
\begin{lstlisting}
Sampled parse 1: condition(pulled_muscle_in_shoulder_in_match({athlete: 'kay', match: 1}) && beat({team1: ['kay'], team2: ['avery'], match: 1}))

Sampled parse 2: condition(pulled_muscle_in_match({athlete: 'kay', match: 1}) && beat({team1: ['kay'], team2: ['avery'], match: 1}))
\end{lstlisting}
\end{tcolorbox}

\paragraph{Retrieving informal relevant background knowledge and proposing conceptual dependency graph}
We jointly sample the informal relevant background information $K$ and corresponding dependency graphs $G$. In our experiments, for each simulated participant we sample $k_{informal}=8$ informal specifications and their corresponding dependency graphs at $temp=0.5$. We then implement an LLM-based evaluation function $\Phi_{informal}$ which jointly scores the generated $K$ and $G$, from which we select the top scoring $K*, G*$. \\

\noindent The full frame prompt for this stage can be found at \texttt{generate-informal-background-knowledge-and-dependency-graph} and the evaluation prompt can be found at \texttt{score-informal-background-and-dependency-graph} in the frame prompts directory. As each generation stage is conditioned on all previous generation steps, note that the injected shuffled and concatenated examples now draw from the input scenario up to the example depdendency graph  (delimited by \texttt{<START\_SCRATCHPAD>}).\\ \clearpage

\noindent Here we show example retrieved informal background knowledge and  the corresponding dependency graph for sample scenarios in \textbf{Exp. 2} (which required retrieving additional information to make up for the underspecified background) and \textbf{Exp. 3}.

\begin{tcolorbox}[title=\textbf{Example informal background knowledge and dependency graph for Exp. 2, biathlon}, colback=white,colframe=white!50!black]
In this event, teams of players are competing in rounds of a biathalon, a winter sport that combines cross-country skiing races and rifle shooting. In each round, the team that wins depends on the average speed with which the athletes are able to ski, based on their intrinsic strength, as well as each team member's shooting accuracy in that particular round.

Intrinsic strength is an underlying attribute of a given athlete that varies somewhat widely from athlete to athlete. An athlete's intrinsic strength ranking out of n other athletes is the number of other athletes we might expect them to be stronger than out of N total random athletes.

Athletes also vary in their shooting accuracy in any given round. Shooting accuracy is a continuous parameter, measured as a percentage from 0 to 100\%. Athletes can have poor, average, or excellent shooting accuracy in a given round. Their shooting accuracy in a round is somewhat dependent on their intrinsic strength, as stronger athletes are probably more likely to have better shooting accuracy.

A team's skiing speed in a round is described in the background as the average effective skiing speed with which the athletes are able to ski on that team. The effective skiing speed of an athlete in a round is determined at a base level by their intrinsic strength, since stronger athletes ski faster.

A team's overall score in a round is the sum of their average skiing speed and their average shooting accuracy in that particular round. The skiing speed and shooting accuracy are weighted equally.

The team that gets the highest score in any given round wins. A team beats another if their team's overall score is higher in that round. A team loses to another team if they did not beat that team in a given round.

The amount that one team wins over another team is given by the difference in their overall scores.
\begin{lstlisting}
- intrinsic_strength
- intrinsic_strength_rank
  - depends on: intrinsic_strength

- shooting_accuracy_in_round
  - depends on: intrinsic_strength

- effective_skiing_speed_in_round
  - depends on: intrinsic_strength

- team_skiing_speed_in_round
  - depends on: effective_skiing_speed_in_round

- team_shooting_accuracy_in_round
  - depends on: shooting_accuracy_in_round

- team_overall_score_in_round
  - depends on: team_skiing_speed_in_round, team_shooting_accuracy_in_round

- beat 
  - depends on: team_overall_score_in_round
- lost
  - depends on: beat
- who_would_win_by_how_much
  - depends on: lost
\end{lstlisting}
\end{tcolorbox}
\clearpage
\begin{tcolorbox}[title=\textbf{Example informal background knowledge and dependency graph for Exp. 3, tug of war}, colback=white,colframe=white!50!black]
\textbf{Participant-generated detail: In the first match of tug-of-war Kay, while managing to pull off the win from Avery, pulled a muscle in their shoulder that limited their pulling output going forward.}\\
First, let's reason about the role of strength in this scenario.
Intrinsic strength is an underlying attribute of a given athlete that varies somewhat widely from athlete to athlete.

An athlete's intrinsic strength ranking out of n other athletes is the number of other athletes we might expect them to be stronger than out of N total random athletes.

Athletes also vary in the effort that they put into any given match. Athletes can put in either moderate amount of effort, little effort, or extra high amounts of effort. Which of these they are more likely to do probably depends on their underlying strength, as stronger athletes are probably more likely to put in extra high effort, and weaker athletes probably tend to be more likely to put in lower amounts of effort.

An athlete who 'tries hard' in a match puts in a fair amount of effort.

Whether or not an athlete pulls a muscle in their shoulder in a specific match occurs at a rare frequency for any given athlete and match. This is an event that affects future matches after the match in which someone was injured. We will need to think about whether an athlete has pulled a muscle in ANY previous matches to understand its effects on the current match.

An athlete's effective pulling strength in a given match is determined at a base level by their intrinsic strength, but is (1) reduced if they pulled a muscle in their shoulder in any PREVIOUS match, which will make them pull less hard; and (2) increased by their effort level, which is effectively a percentage multiplier on their pulling strength in this match.

A team's pulling strength in a match is described in the background as the AVERAGE effective pulling strength with which the athletes are able to pull on that team.

A tug-of-war team beats another if their team's pulling strength is greater in that match, assuming a fixed match length.

A tug-of-war team loses to another team if they did not beat that team in a given match.

To calculate who would win and by how much, we will calculate the likelihood that a team would win over another.
\begin{lstlisting}
- intrinsic_strength
- intrinsic_strength_rank
  - depends on: intrinsic_strength
- effort_level_in_match
  - depends on: intrinsic_strength
- pulled_muscle_in_shoulder_in_match
- pulled_muscle_in_shoulder_in_any_previous_match
  - depends on: pulled_muscle_in_shoulder_in_match
- effective_athlete_pulling_strength_in_match
  - depends on: intrinsic_strength, pulled_muscle_in_shoulder_in_any_previous_match, effort_level_in_match
- team_pulling_strength_in_match
  - depends on: effective_athlete_pulling_strength_in_match
- beat 
  - depends on: team_pulling_strength_in_match
- lost
  - depends on: beat
- who_would_win_by_how_much
  - depends on: lost
\end{lstlisting}
\end{tcolorbox}

\paragraph{Synthesizing the formal symbolic model}
Finally, conditioned on the best scoring parse, informal knowledge and dependency graph, for each simulated participant we sample a single full probabilistic program $k_{program}=1$ at $temp=0.2$ from the LLM. As described in the main text, we implement $\Phi_{formal}$ as simply whether or not the probabilistic program compiles and produces inferences over the queries.\\

\noindent The full frame prompt for this stage can be found at \texttt{generate-model} in the frame prompts directory. As each generation stage is conditioned on all previous generation steps, note that the injected shuffled and concatenated examples now include the full text of the examples, from the scenarios and including all prior example generation stages up to the example models.\\ 
As models are verbose, below we reproduce an \textit{excerpted} model from \textbf{Exp. 3} showing the parse that involves functions for the new participant detail, and with excerpted components for the other named functions.
\begin{tcolorbox}[title=\textbf{Example informal background knowledge and dependency graph for Exp. 3, biathlon}, colback=white,colframe=white!50!black]
\textbf{Participant-generated detail: In the first match of tug-of-war Kay, while managing to pull off the win from Avery, pulled a muscle in their shoulder that limited their pulling output going forward.}\\
\begin{lstlisting}
var model = function() {
// BACKGROUND KNOWLEDGE

    // Intrinsic strength is an underlying attribute of a given athlete that varies somewhat widely from athlete to athlete.
    var intrinsic_strength = mem(function ({athlete}) {...})

    // An athlete's intrinsic strength ranking out of n other athletes is the number of other athletes we might expect them to be stronger than out of N total random athletes.
    var intrinsic_strength_rank = function({athlete, out_of_n_athletes}) {...}

    // Athletes also vary in the effort that they put into any given match. Athletes can put in either moderate amount of effort, little effort, or extra high amounts of effort. Which of these they are more likely to do probably depends on their underlying strength, as stronger athletes are probably more likely to put in extra high effort, and weaker athletes probably tend to be more likely to put in lower amounts of effort.
    var effort_level_in_match = mem(function({athlete, match}) {...})
    
    // An athlete who 'tries hard' in a match puts in a fair amount of effort.
    var tried_hard_in_match = mem(function({athlete, match}) {...})

    // Whether or not an athlete pulls a muscle in their shoulder in a specific match occurs at a rare frequency for any given athlete and match. This is an event that affects future matches after the match in which someone was injured. We will need to think about whether an athlete has pulled a muscle in ANY previous matches to understand its effects on the current match.
    var pulled_muscle_in_shoulder_in_match = mem(function({athlete, match}) {
      var likelihood_of_pulling_muscle_in_match = 0.05;
      return flip(likelihood_of_pulling_muscle_in_match); 
    })

    // An athlete's effective pulling strength in a given match is determined at a base level by their intrinsic strength, but is (1) reduced if they pulled a muscle in their shoulder in any PREVIOUS match, which will make them pull less hard; and (2) increased by their effort level, which is effectively a percentage multiplier on their pulling strength in this match.
    var effective_athlete_pulling_strength_in_match = mem(function({athlete, match}) {
      // Assume that base pulling strength is just their current strength.
      var base_pulling_strength_in_match = intrinsic_strength({athlete : athlete})

      //  Reduced if they pulled a muscle in their shoulder in any PREVIOUS match. Use the helper function to check if they pulled a muscle after any previous match.
      var pulling_strength_adjusted_for_pulled_muscle = any_previous_time_inclusive(
        function(prev_match) {
          return pulled_muscle_in_shoulder_in_match({athlete: athlete, match: prev_match})
        }, match) ? base_pulling_strength_in_match * 0.7 : base_pulling_strength_in_match;

      // Increased by effort level in this match.
      var pulling_strength_adjusted_for_effort_level = (effort_level_in_match({athlete: athlete, match: match}) / 100) * pulling_strength_adjusted_for_pulled_muscle;

      return pulling_strength_adjusted_for_effort_level; 
    })

    // A team's pulling strength in a match is described in the background as the AVERAGE effective pulling strength with which the athletes are able to pull on that team.
    var team_pulling_strength_in_match = function({team, match}) {...}
    
    // A tug-of-war team beats another if their team's pulling strength is greater in that match, assuming a fixed match length.
    var beat = function({team1, team2, match}){...}
    
    // A tug-of-war team loses to another team if they did not beat that team in a given match.
    var lost = function({team1, team2, match}){...}

    // To calculate who would win and by how much, we will calculate the likelihood that a team would win over another. 
    var who_would_win_by_how_much = function({team1, team2, match}) {...}
    
    <CONDITION AND QUERIES OMITTED FOR CONCISION>
    }}
var posterior = Infer({ model: model, method: 'rejection'});
\end{lstlisting}
\end{tcolorbox}

\paragraph{Model-based Bayesian inferences}
In our experiment, all inferences are derived using the WebPPL built in \textit{rejection sampling} inference engine. Inference budgets are specified in the main text: we report posteriors from $b_{samples}=1000$ samples per simulated participant for \textbf{Exp. 1} and \textbf{Exp. 2}, and $b_{samples}=500$ for \textbf{Exp. 3} (as in general rejection sampling is much slower on these, where observations specify a rare a priori observation.

\subsection{Natural Language Reasoning Experiments: Additional Experimenstal Details}
\paragraph{Model Olympics Vignettes}
\noindent This supplemental section provides additional details on the stimuli generation and selection process for the Model Olympics domain vignettes used throughout the experiments. \\

\noindent As described in the main text, we construct a set of procedurally generated vignettes for experiments \textbf{Exp. 1, Exp. 2, Exp. 3}, where each vignette consists of a: linguistic \textit{background} on the particular sport of interest (which could be \textit{tug-of-war}, \textit{canoe racing}, or \textit{biathlon}); a set of \textit{evidence} sentences describing match outcomes (plus, in the \textbf{Exp. 3} case, one additional participant-generated observation); and 8 \textit{questions}. \\\\
At the data repository section, the \texttt{model-olympics-human-experiment} directory contains:
\begin{itemize}
\itemsep0em 
\item Base \textbf{detailed backgrounds} for the \{\texttt{tug-of-war, canoe-racing, biathlon}\} sports used for vignettes in \textbf{Exp. 1}.
\item Base \textbf{underspecified backgrounds} for the \{\texttt{tug-of-war, canoe-racing, biathlon}\} sports used for vignettes in \textbf{Exp. 2}.
\item Base \textbf{underspecified backgrounds (no reference to any participant-generated variables}) for the \{\texttt{tug-of-war, canoe-racing}\}  sports used for vignettes in \textbf{Exp. 3}. Note that the Exp. 3 vignettes were constructed shown to models were constructed using underspecified backgrounds (as in Exp. 2); we provide these again for comparison.
\end{itemize}

\noindent Using the base backgrounds, we procedurally generated vignettes for each sport using a set of \textbf{16 base vignette templates}, comprised of \textbf{12} templates derived from the patterns of evidence used (originally, in the tug-of-war domain only) in \cite{goodman2014concepts}, and \textbf{4} additional templates specifically designed to present noisy and anomalous evidence that would evaluate whether participants and models judged these outcomes based on the ``Bayesian explaining away" of anomalous outcomes relative to accumulative contrary evidence, based on multiple conjunctive latent causal variables. The templates describe the relations between athletes in a tournament; we instantiate the templates into concrete templates for each sport using sport-specific latent variables, and with randomly sampled athlete names from a set of gender-neutral names (to avoid priors about athlete strength). We then randomly subsampled amongst these procedurally templates to select the vignettes reported in our experiments. In total, as described in the main experimental text the final stimuli for each experiment reported in the paper comprised:
\begin{itemize}
\itemsep0em 
\item \textbf{Exp. 1}: 6 randomly sampled vignettes (from the full  set of 16 possible vignette templates) for each sport, for a total of \textbf{18} vignettes. Note that these 6 vignette templates were independently sampled for each sport and therefore may not have had the same evidence patterns per sport.
\item \textbf{Exp. 2:} matched vignette templates to Exp. 1, for a total of \textbf{18} vignettes, but with underspecified backgrounds and re-generated athlete names.
\item \textbf{Exp. 3:} 5 tug-of-war and 4 canoe-racing vignettes, which were base vignettes extended with participant-generated details. As we describe throughout, these base vignettes were similar in form but slightly different (in their background details and phrasing of the inference questions) from those used in Exp. 1 and Exp. 2, as this was a preliminary experiment piloted before Exp. 1 and Exp. 2. 
\end{itemize}
As we describe in the human experimental details below, human participants in our study actually viewed slightly more vignettes than were ultimately reported in our paper here or compared to model results -- we removed one vignette (from all three reports) which contained an error in the questions that asked about an athlete who was actually not part of a particular match; and we removed one additional sports domain, a synchronized\textit{diving} domain, due to apparent confusion about the sport itself and high amounts of variance in participant answers. We currently withhold the full exact set of stimuli from \textbf{Exp. 1} and \textbf{Exp. 2}, and the stimuli used in \textbf{Exp. 3}, to avoid their appearance in LLM training datasets while we prepare an extended version of this work. The full dataset will be released upon publication, and future work will seek to generate a more dynamic version of this dataset for evaluation. However, here we show an example vignette from the \textit{tug-of-war} domain, demonstrating the difference between the detailed (Exp. 1) and underspecified (Exp. 2) backgrounds.

\begin{tcolorbox}[title=\textbf{Example tug-of-war vignette, showing Exp. 1 vs. Exp. 2 backgrounds}, colback=white,colframe=white!50!black]
\textbf{Exp. 1: Detailed background}
In this event, the athletes are competing in tug-of-war tournaments. Each tournament consists of a series of matches. In each match, athletes compete as part of a team. \\
An athlete’s intrinsic strength remains constant throughout a tournament. An athlete neither gets stronger nor weaker between matches. You can assume that all matches take place on the same day.\\
Athletes also vary in the effort that they put into any given match. Most of the time, people pull with a moderately high amount of effort. Sometimes, an athlete won’t put in much effort and will pull with only a fraction of their strength. Other times, they may put in a lot of effort and pull extra hard, beyond what their intrinsic strength would suggest.\\
How hard a team pulls overall in any given match is determined by the total amount that all of the athletes on the team pull in that match. How hard each athlete pulls in a given match is determined by their intrinsic strength, modified by how much effort they put in (a lower fraction of their intrinsic strength if they don’t put in much effort, or even more than their strength if they put in more effort).\\
The team that pulls the hardest in a given match wins.\\
Athletes compete either individually or as a team.\\
All matches take place on the same day.\\

\textbf{Exp. 2: Underspecified background}
In this event, the athletes are competing in matches of tug-of-war.\\
In each round, the team that wins the round depends on how hard the athletes collectively pull, based on their intrinsic strength modulated by other factors including how much effort they put in to that round.\\
Athletes compete either individually or as a team.\\
All matches take place on the same day.\\\\

\textbf{CONDITIONS}\\
In the first match, Peyton and Avery lost to Blake and Casey.\\
In the second match, Peyton and Blake lost to Avery and Casey.\\
In the third match, Peyton and Casey lost to Avery and Blake.\\

\textbf{QUERIES}\\
Query 1: Out of 100 random athletes, where do you think Peyton ranks in terms of intrinsic strength?\\
Query 2: Out of 100 random athletes, where do you think Avery ranks in terms of intrinsic strength?\\
Query 3: Out of 100 random athletes, where do you think Blake ranks in terms of intrinsic strength?\\
Query 4: On a percentage scale from 0 to 100\%, how much effort do you think Peyton put into the second match?\\
Query 5: On a percentage scale from 0 to 100\%, how much effort do you think Avery put into the second match?\\
Query 6: On a percentage scale from 0 to 100\%, how much effort do you think Blake put into the second match?\\
Query 7: In a new match later this same day between Peyton and Avery (Team 1) and Blake and Gale (Team 2), who would win and by how much?\\
Query 8: In a new match later this same day between Peyton and Blake (Team 1) and Avery and Gale (Team 2), who would win and by how much?\\
\end{tcolorbox}

\paragraph{LM-only experimental details} The repository includes the frame prompting format used to elicit judgments for both the \textit{LM-direct} and \textit{LM-CoT} baselines, in the \texttt{lm-only-baseline-prompts} directory.

Each prompt contained the full \textit{experiment instructions} shown to humans for each experiment (though note that the videos showing how to use the multi-click judgment interface were described in text, as prompts were text only); and then the full vignette, with additional instructions for how to answer each query.


\begin{figure*}
    \centering
    \includegraphics[width=0.8\linewidth]{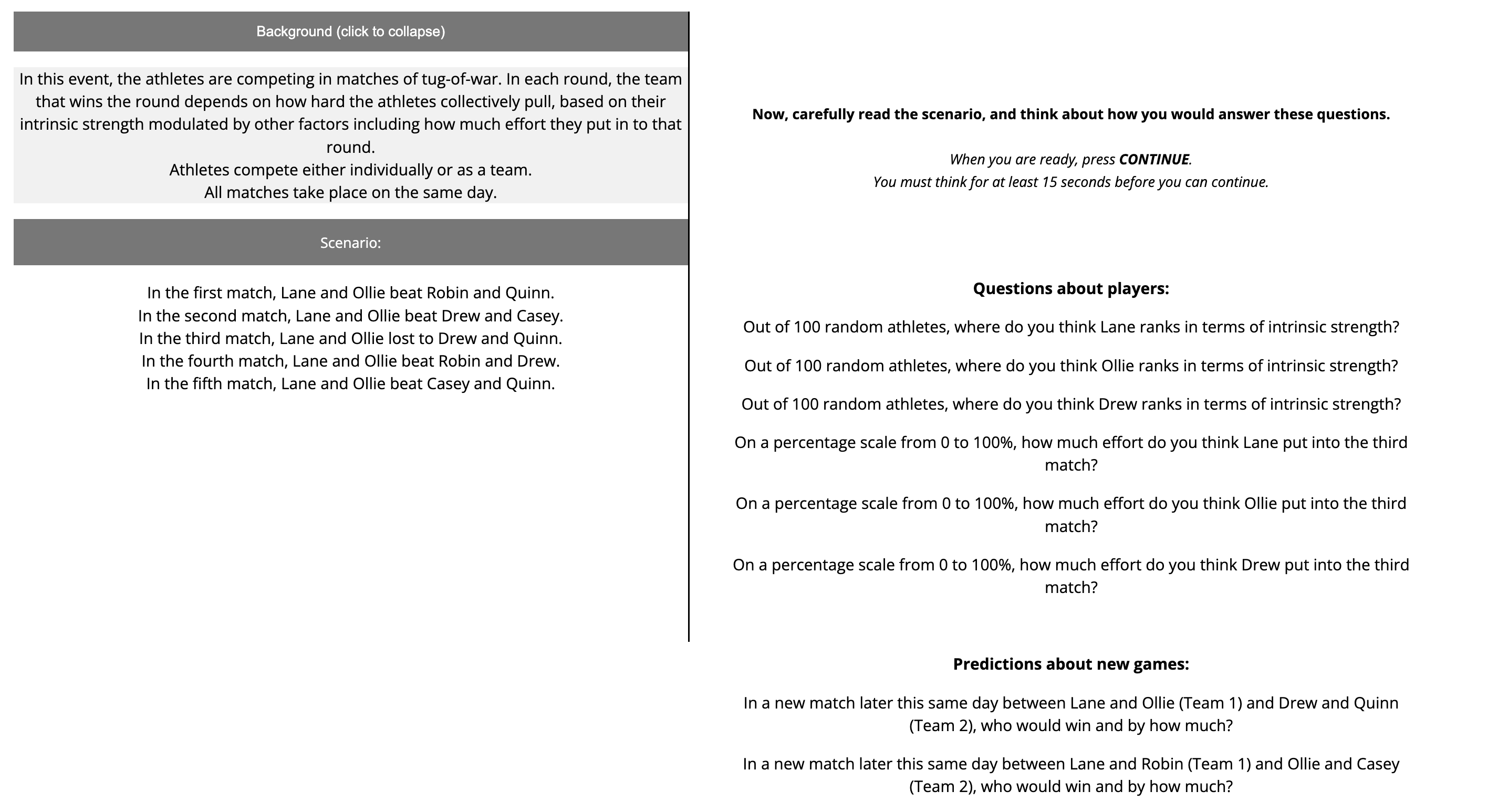}
    \rule{\linewidth}{0.4pt} 
    \includegraphics[width=0.8\linewidth]{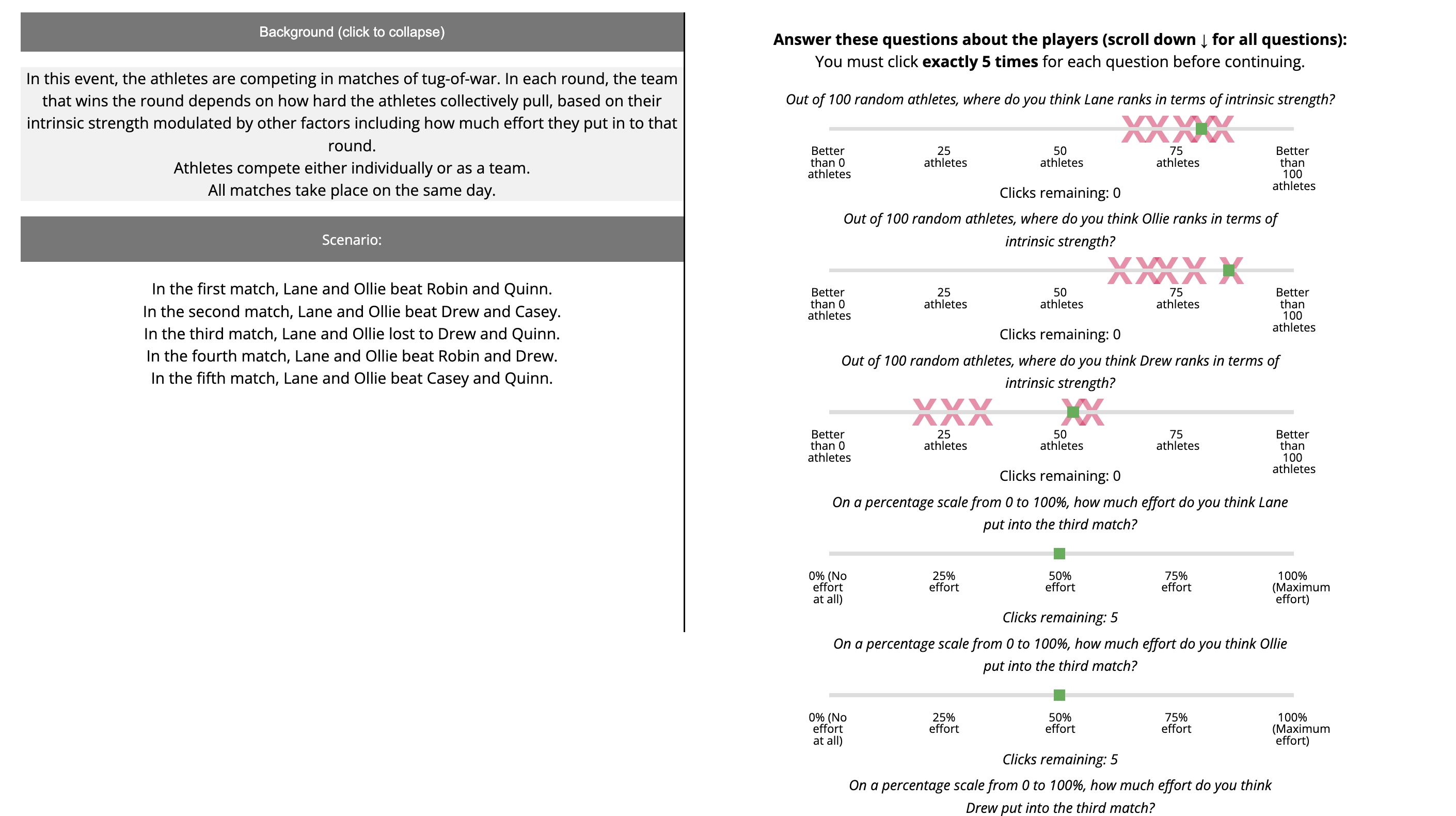}
    \caption{Example interfaces showing the human experimental setup -- shown is a sample trial from \textbf{Exp. 2}, the underspecified background experiment. Participants first read the background information, scenario and questions (top). They then indicate their judgments via multiple clicks per question (bottom)}
    \label{fig:interface}
\end{figure*}

\paragraph{Exp. 1: human judgment experimental details} Pre-trial instructions were shown to all participants containing an example tutorial on how to use the multi-click slider interface to indicate the distribution of their judgments, including GIFs showing how they could indicate high certainty about a specific posterior mode (eg. most clicks around one end of the slider); split certainty about multiple posterior modes; and relative uncertainty about a continuous range. 

As described in the main text, participants judged a randomly constructed batch of two vignettes from each of the three sport. Trials were grouped by sport  (all participants first saw vignettes about tug-of-war, then canoe-races, then biathlon. Participants in $3$ of the $4$ batch conditions ($57$ of $76$ participants) also saw an additional 2 vignettes from an additional synchronized diving domain; this domain was moved after participants appeared generally confused about the domain and showed extremely low inter-participant correlation in answers.

Trials were grouped into sections by each sport. Prior to reading the vignettes for a particular sport, in Exp. 1 only, participants were additionally presented with a full background description of the sport (with the same details as in the \textit{detailed background} D(, along with an example tournament showing the kinds of outcome patterns that could appear in the later vignettes). Participants were required to spend 15 seconds reading this background description. The full text of these background descriptions can be found in the \texttt{model-olympics-human-experiment} directory of our repository. Then, for each vignette trial, participants first read were (1) presented with the background and vignette containing evidence and questions (but no sliders for inputs), as shown in  the example interface in \autoref{fig:interface} (\textbf{top}), and required to think about the vignette for 15 seconds without progressing; they coudl then could proceed to (2) an interface presenting sliders for the multi-click inputs, shown in \autoref{fig:interface} (\textbf{bottom}). Participants took a median time of $2.24$ minutes to provide all judgments for one vignette. Participants were paid at a base rate of \$15/hr and told they may receive a bonus of up to \$16/hr ``if you try your best throughout the experiment to answer each question''; in reality, all participants were provided the bonus.  

\paragraph{Exp. 2: human judgment experimental details} This experiment followed the same interface format as Exp. 1, except with the underspecified backgrounds for each vignette. Additionally, participants in Exp. 2 were not shown the additional full description of the sport background prior to beginning the vignettes (they only read the backgrounds alongside the vignettes themselves).  Participants took on average $2.81$ minutes to provide their multi-click judgments per vignette. As with Exp. 1, participants in $2$ of the $4$ sets of vignettes also saw vignettes about diving, along with the other three sports, which were later omitted from the experimental analysis when this sport was removed from analysis.

\paragraph{Exp. 3: human judgment experimental details} This experiment involved both a \textbf{human commentary elicitation} experiment and a \textbf{human judgment experiment}.

During the \textbf{commentary elicitation experiment}, as described in the main text, N=20 participants were shown a tutorial indicating that they would read vignettes about sports scenario, and then ``act as a sports commentator" to write one or a few sentences introducing a new detail that would \textit{change their} reasoning about a randomly selected new match prediction question. Participants were randomly assigned to conditions over which of the two new match questions they would need to change, and whether they were asked to produce details that would either \textit{increase} or \textit{decrease} the odds of a particular outcome given their initial judgments. On each trial, as in earlier experiments, participants completed the full judgment task -- they read the vignettes for 15 seconds, then proceeded to the sliders where they entered judgments for all questions. They were then shown which new match prediction they were to manipulate and told the direction they would need to manipulate with their commentary. After writing commentary, participants were shown the full vignette (with their added commentary) and asked to re-enter their judgments on the new match prediction.

As noted in the main text, this experiment used only the tug-of-war and canoe racing sports; and used 9 base vignettes with the \textit{underspecified} backgrounds from Exp. 2, and slightly different patterns of evidence than used in Exp. 1 and Exp. 2 -- in particular, the vignettes included slightly easier outcome patterns of evidence involving head-head matches between single players, whereas the vignettes in Exp. 1 and Exp. 2 only involved matches between teams of two players each. Additionally, participants were shown slightly different wordings of the \textit{judgment} questions during the trials: the strength questions asked \textit{how strong} athletes were (rather than their absolute strength ranking out of random athletes) and \textit{how much effort} they put in (rather than asking specifically for a percent effort). 

In total, this experiment yielded an initial set of \textit{81} initial distinct commentary observations across all participants. We filtered these down to 9 final vignettes by (1) excluding all participants who did not adjust their judgments after providing commentary in the specified direction (e.g., they did not actually increase the odds of the predicted match); (2) excluding participants who appeared to have used language models (participants were explicitly instructed not to) or who provided clearly spam answers; (3) excluding commentary that was more than a single sentence. We then selected the 9 commentary with a more specific set of criteria that could be generalized in future work -- we selected commentary that focused on a \textit{single athlete} (rather than generics about the world, like \textit{it was raining}); and commentary that focused on a single new \textit{event} observation (eg. \textit{Athlete A took an energy drink}) or observation about the athlete (\textit{Athlete A had less experience}). 

During the \textbf{human judgment experiment}, we then recruit a new set of participants to provide the same $k=5$ multi-click judgments as in Experiments 1 and 2. Instructions were the same as in Exp. 1 and Exp. 2, except that participants were told that they would be reading vignettes including commentary written by other people. Each participant in this trial was shown the full set of k=9 vignettes with commentary. Participants were provided the \textit{underspecified} sport description from Exp. 2, as described in the main text. Participants took approximately $2.22$ minutes to provide their judgments per vignette.

\subsection{Results: Supplemental Analyses}
This section collects additional analyses comparing human and model judgments.

\begin{figure*}[h!]
    \centering
    \includegraphics[width=0.7\linewidth]{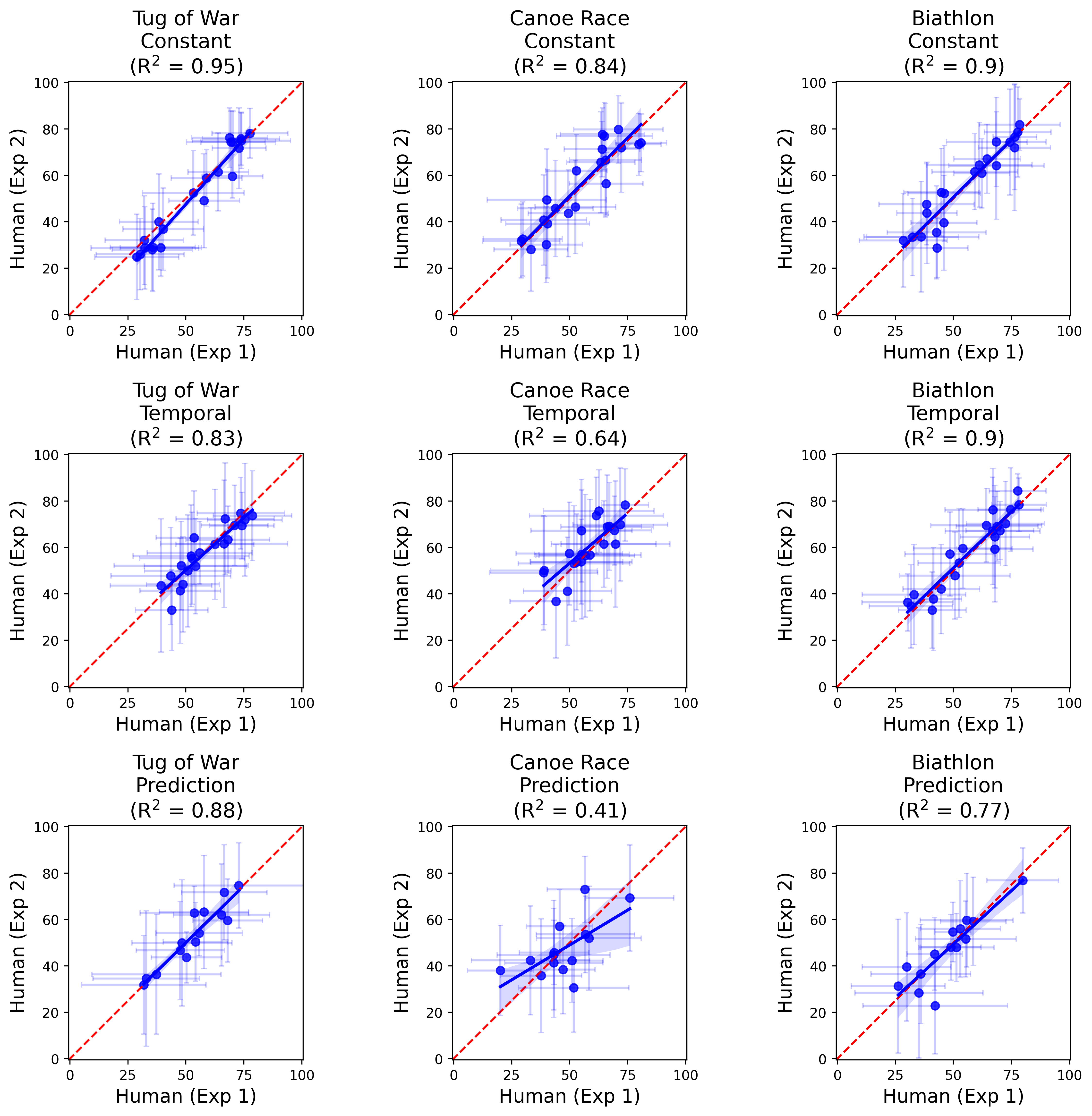}
    \caption{Correlations between human participant predictions per stimuli per query between Experiment 1 (x axis) versus Experiment 2 (y axis).}
    \label{fig:human-compare-e1e2}
\end{figure*}

\paragraph{Human and MSA correlations between Experiments 1 and 2} We first examine how well \textit{human judgments correlate across the matched vignettes in experiment 1 and experiment 2} -- that is, whether people made judgments when reading the detailed background information that correlated with those from the underspecified background information. In general, as seen in \autoref{fig:human-compare-e1e2}, judgments appear to be highly correlated, providing some evidence that people retrieve and use similar kinds of information to reason about the underspecified Exp. 2 condition as those that were provided to them explicitly in Exp. 1. Notably, there appears to be less correlation in the canoe race sport (middle column) -- suggesting that people generally retrieved other ways that effort and strength might have contributed to the observed outcomes when left to come up with these details on their own, compared to the version spelled out to them in Exp. 1. 

\begin{figure*} [h!]
    \centering
    \includegraphics[width=0.7\linewidth]{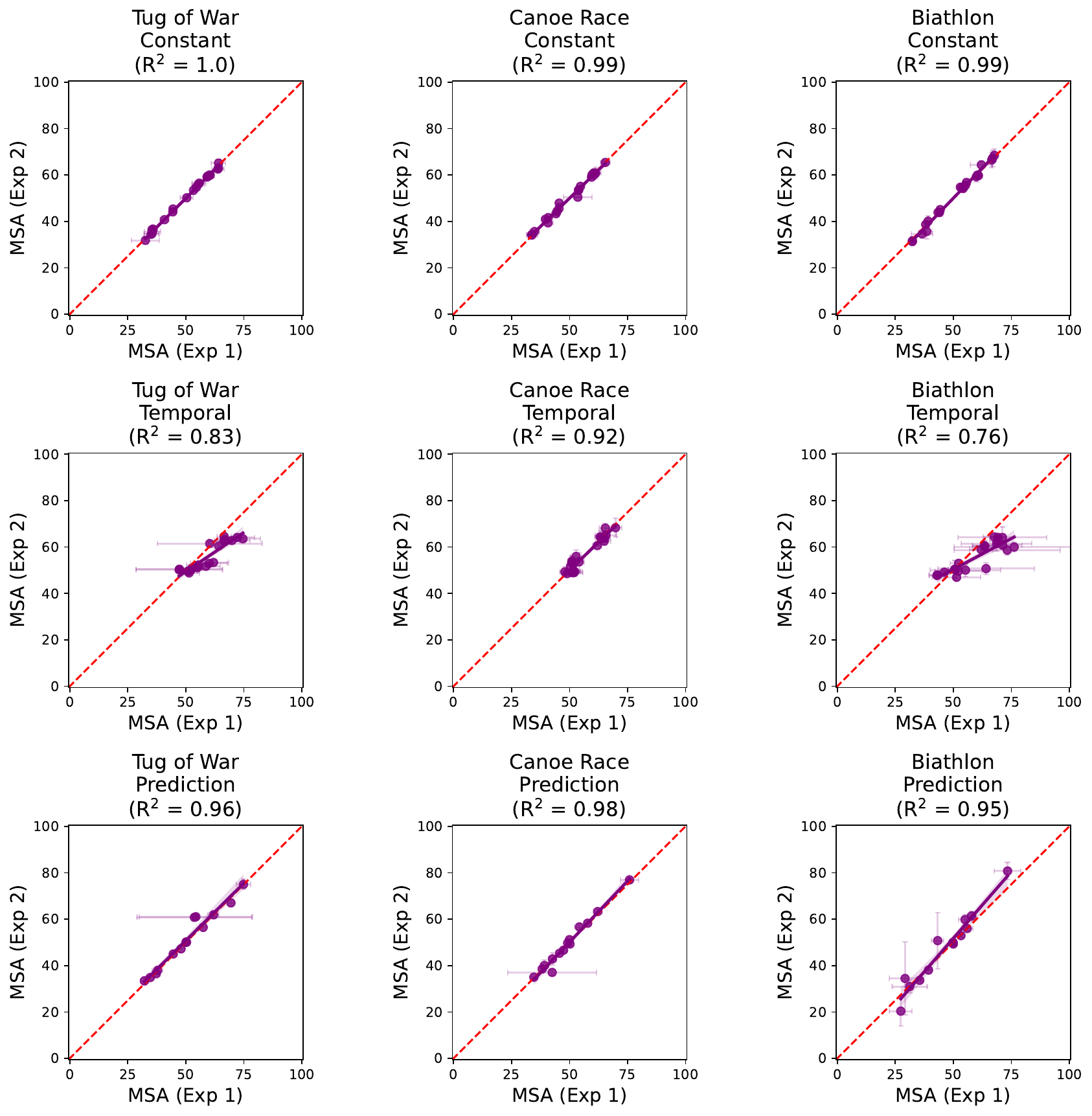}
    \caption{Comparing correlations between MSA predictions per stimuli per query for Experiment 1 (x axis) versus Experiment 2 (y axis).}
    \label{fig:msa-compare-e1e2}
\end{figure*}

We perform the same analysis for the MSA judgments, comparing correlations between MSA judgments in Experiment 1 and 2 (\autoref{fig:msa-compare-e1e2}). In general, we see that the judgments are \textit{very} well correlated between the experiments, more so than the human participants -- and unlike human participants, we do not see variance in the canoe racing condition between Experiment 1 and 2. This warrants further investigation, as it suggests that the model synthesis procedure is less diverse in producing human-like distributions over possible ad-hoc \textit{models} in Experiment 2 from underspecified backgrounds, and may reflect a lack of sampling diversity in model construction.

\begin{figure*} [h!]
    \centering
    \includegraphics[width=1.0\linewidth]{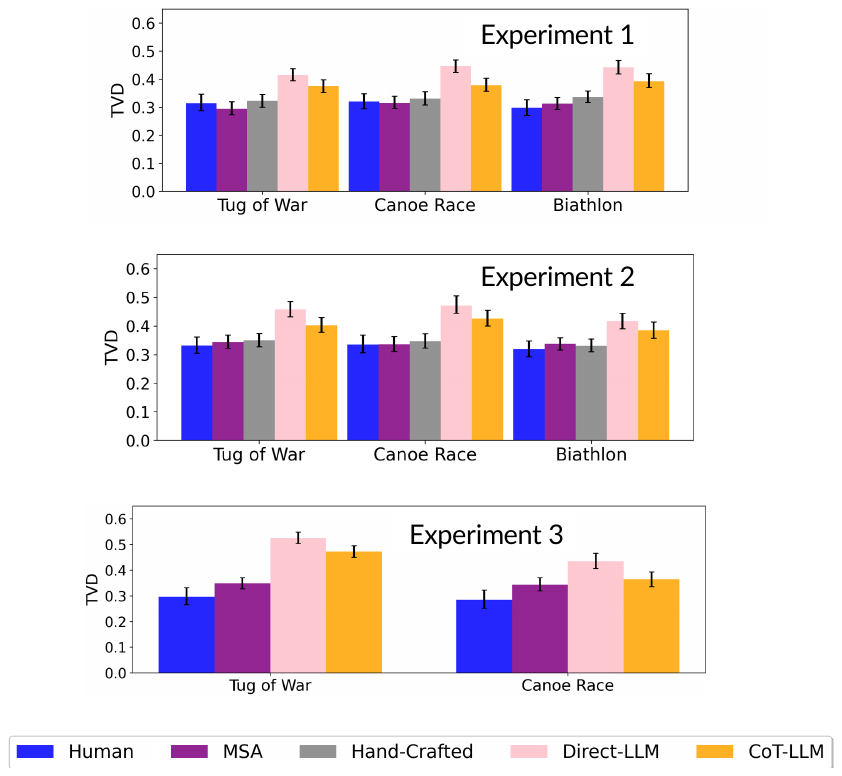}
    \vspace{-3cm}
    \caption{Comparing Total Variation Distance (TVD) between model and human judgments across each experiment. Bootstrapping and averaging follow as in our Wasserstein Distance computations; that is: TVD is computed between judgments per query per scenario, then aggregated as the mean over query types, and mean across query types for each depicted sport and experiment.) Error bars for model-humans show 95\% CI over $1000$ bootstrapped samples, with replacement, on the human data; for human-humans, over $1000$ sampled 50-50 split-halve TVDs.}
    \label{fig:tvd}
\end{figure*}

\paragraph{Total Variation Distance for comparing distributions between humans and models} To ensure our distributional analyses are not specific to using the Wasserstein Distance metric, we repeat our distributional analyses using Total Variation Distance (which does not account for the ``geography'' of the domain when comparing distributions). As in our Wasserstein Distance analyses, we first bucketize participant and model judgments (into $10$ buckets) and compute our measure over the buckets. We see similar trends across models, sports, and experiments in Figure~\ref{fig:tvd}. 

\begin{figure*}[h!]
    \centering
    \includegraphics[width=0.45\linewidth]{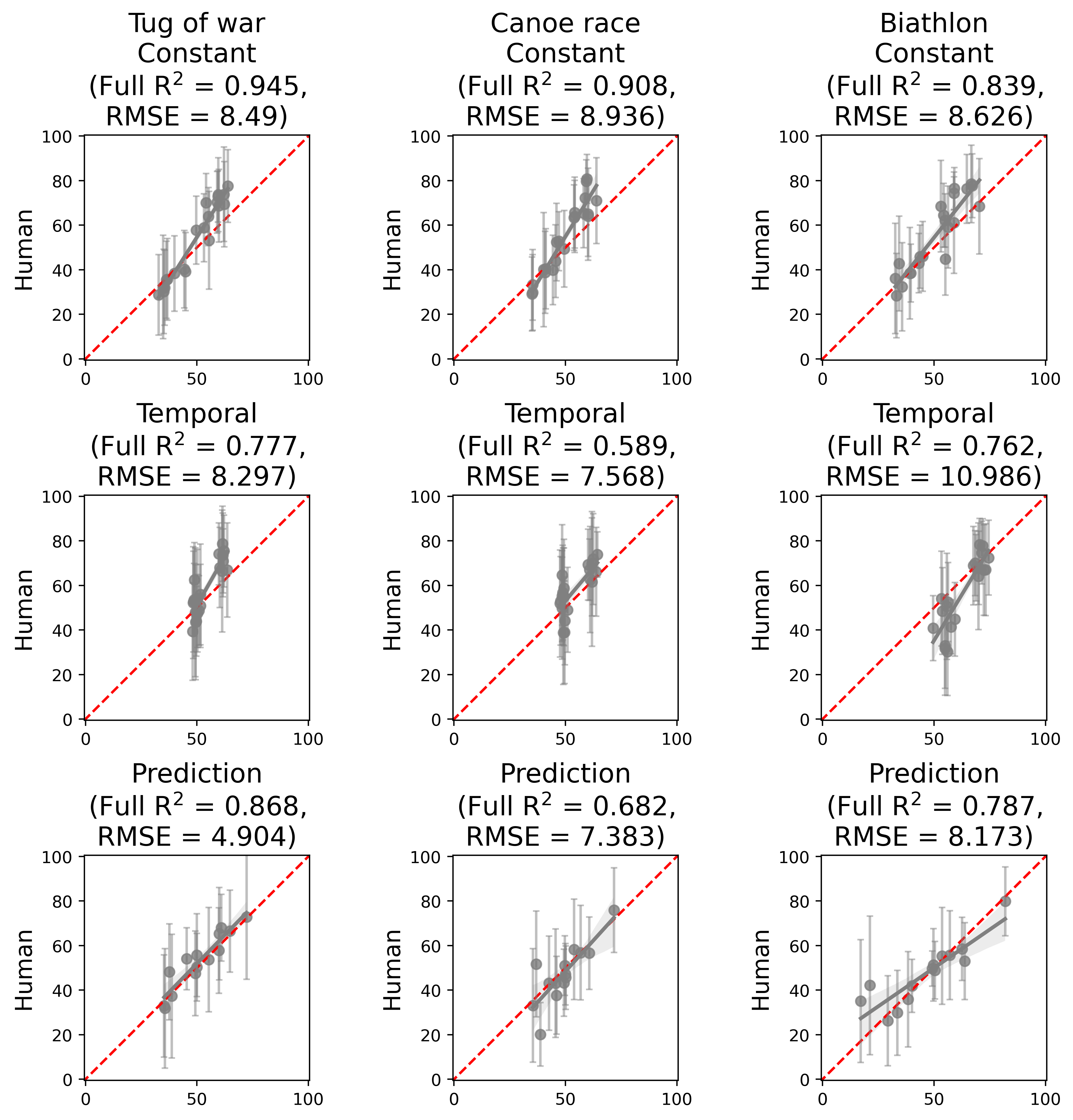}
    \includegraphics[width=0.45\linewidth]{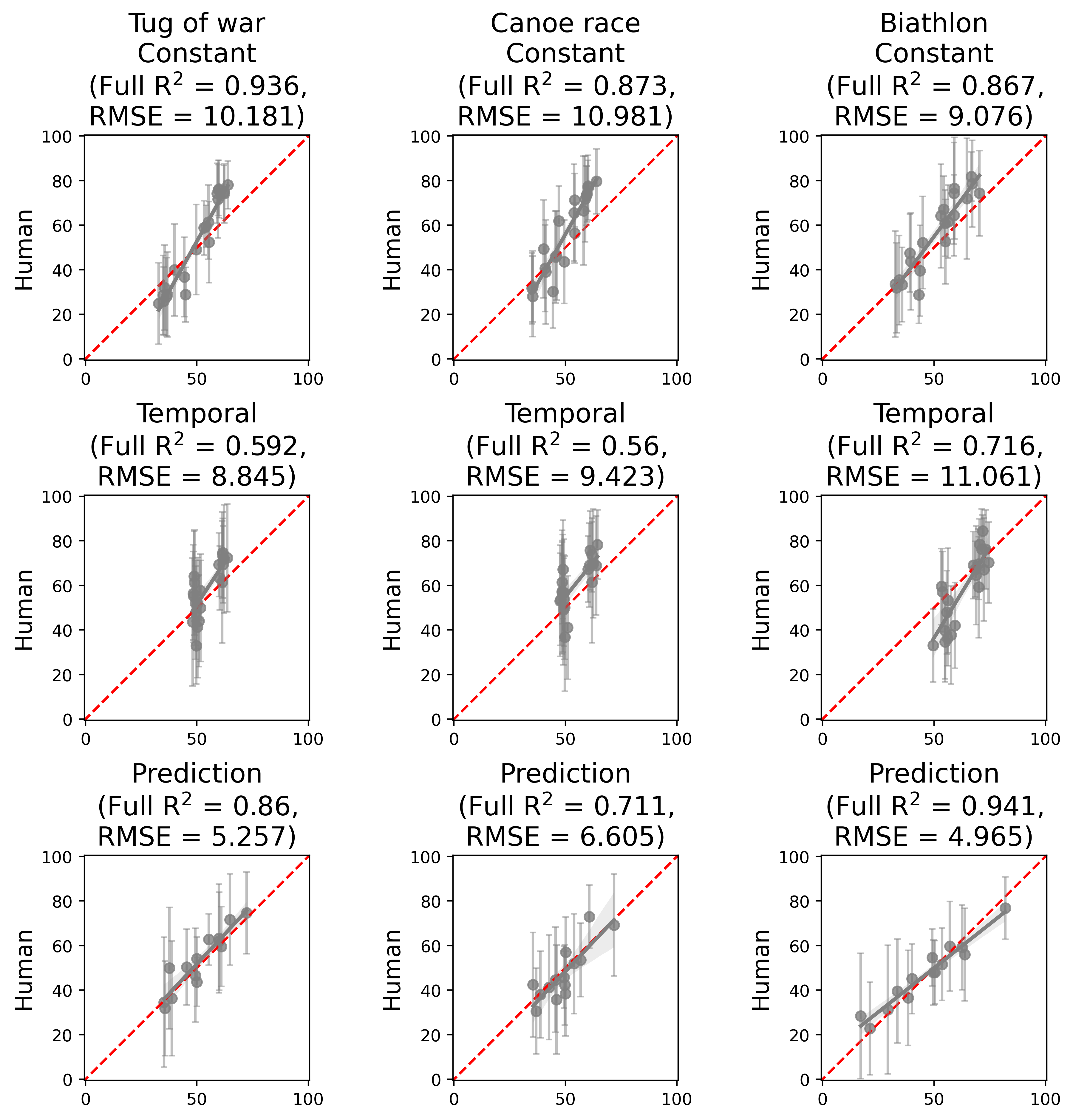}
    \caption{Inferences under the gold model against people for Exp. 1 (left) and Exp. 2 (right). Error bars depict standard deviation over the human responses.}
    \label{fig:e1e2-gold-scatter}
\end{figure*}

\begin{figure*}[h!]
    \centering
    \includegraphics[width=0.45\linewidth]{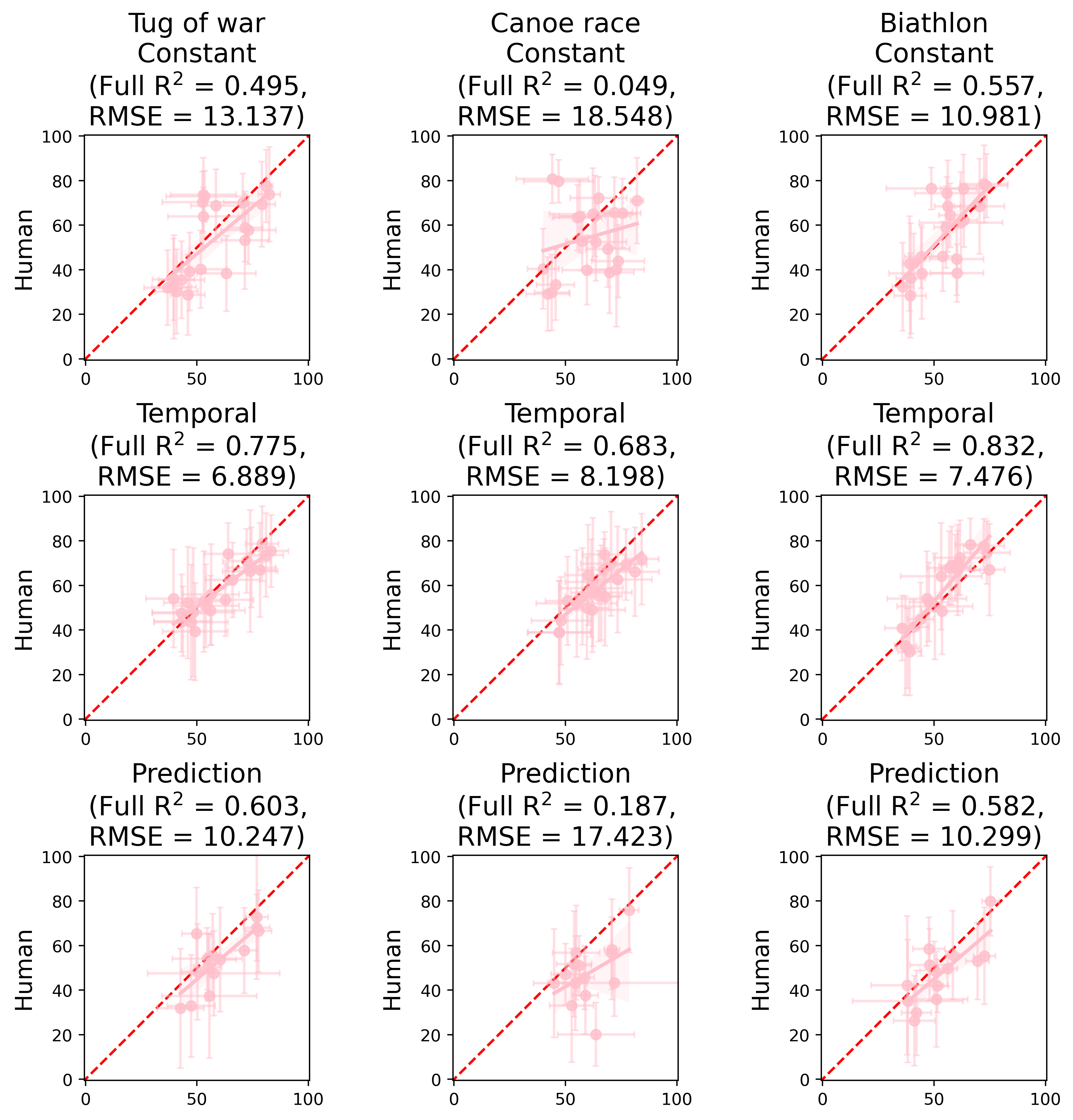}
    \includegraphics[width=0.45\linewidth]{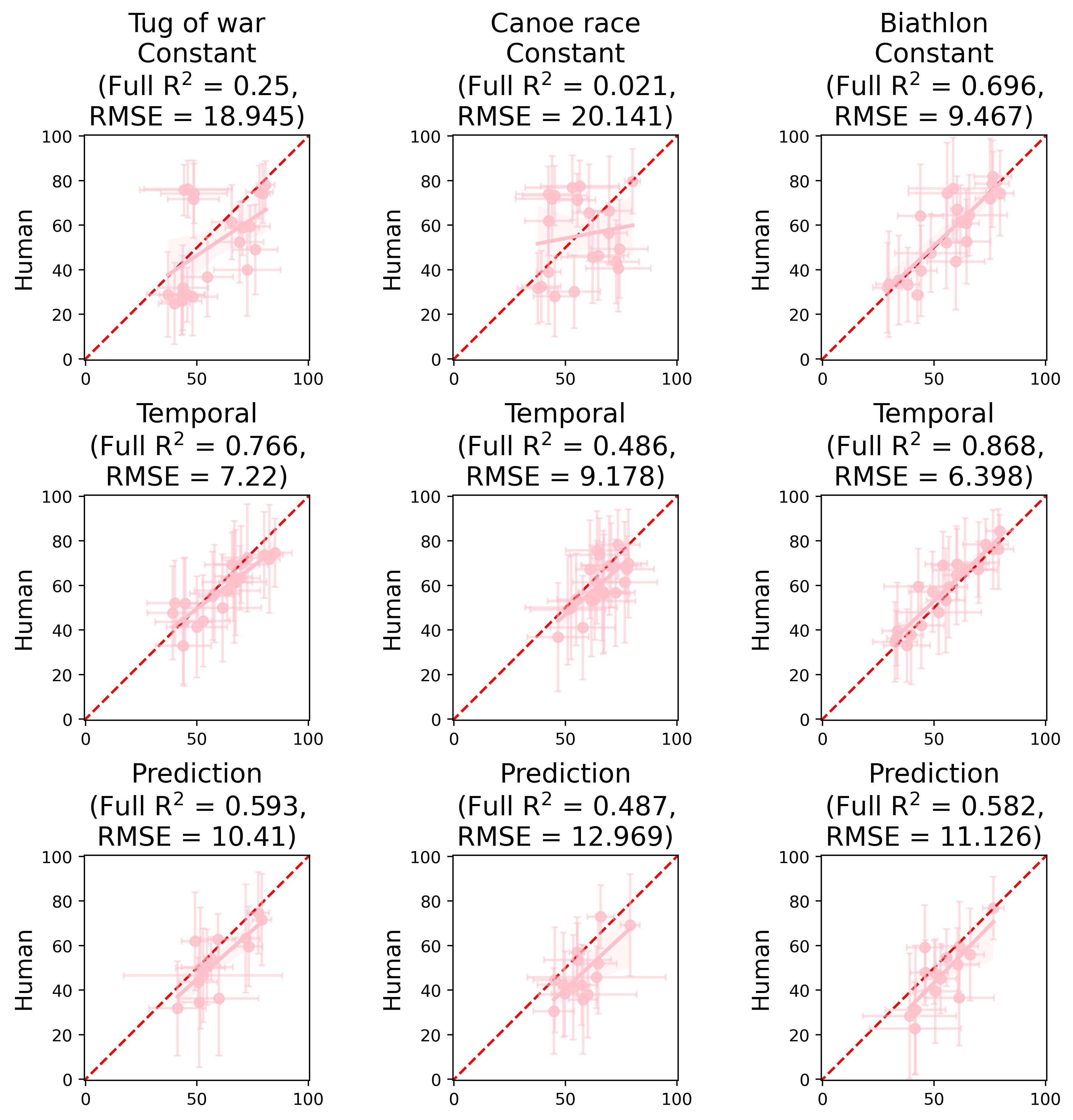}
    \caption{Inferences under the Direct-LLM model against people for Exp. 1 (left) and Exp. 2 (right). Error bars depict standard deviation over the human and model responses.}
    \label{fig:e1e2-vanilla-scatter}
\end{figure*}

\begin{figure*}[h!]
    \centering
    \includegraphics[width=0.45\linewidth]{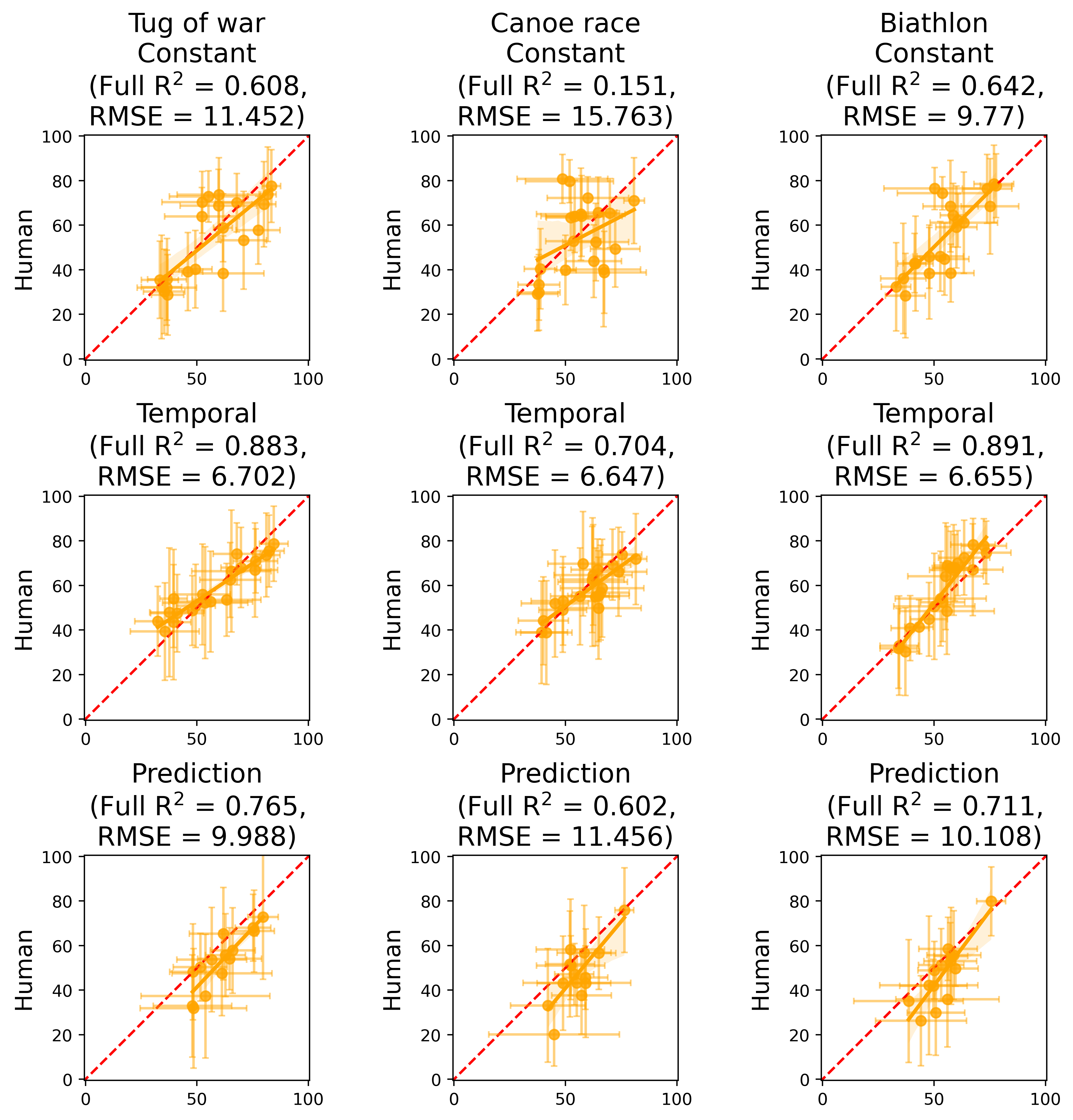}
    \includegraphics[width=0.45\linewidth]{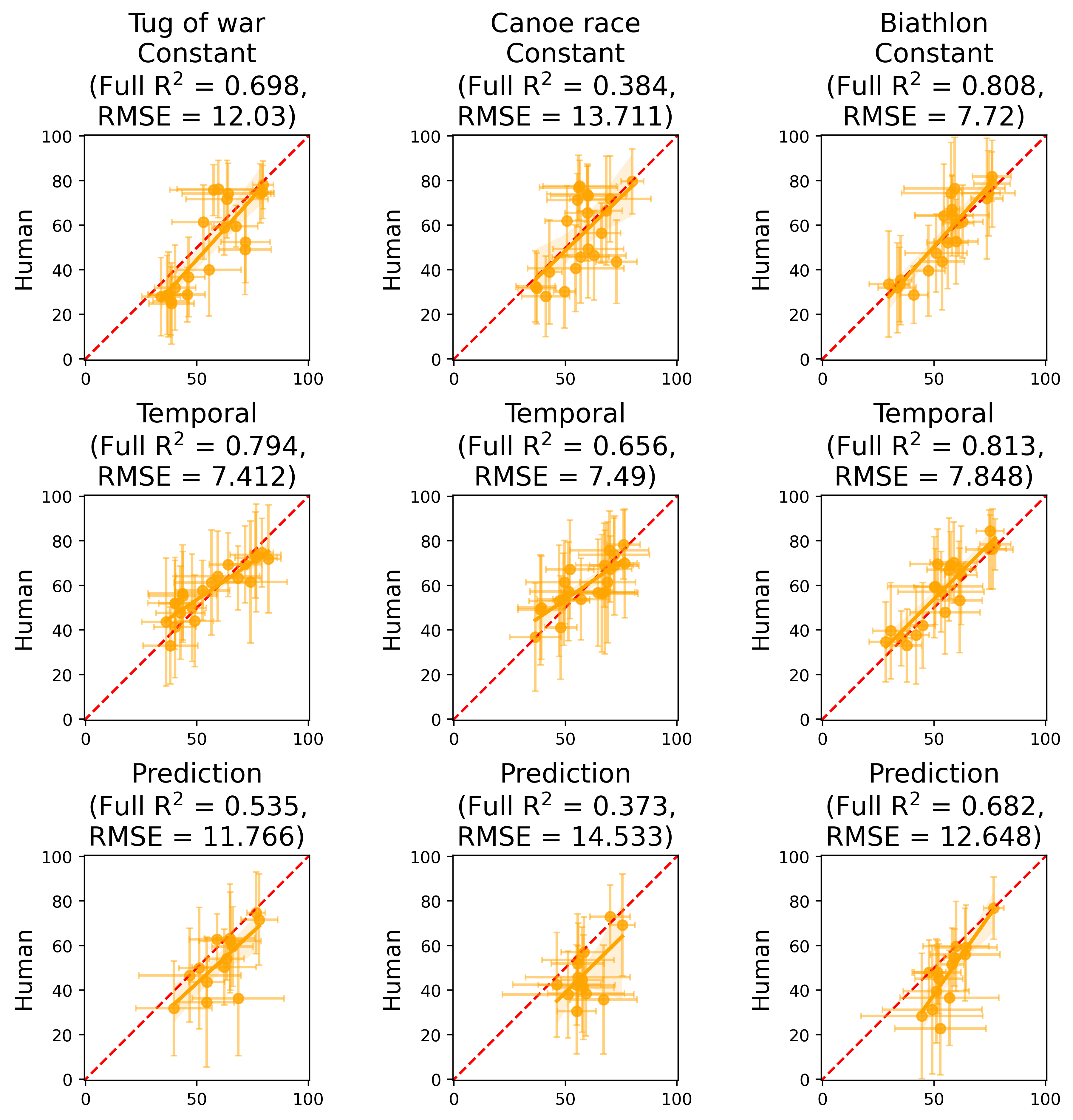}
    \caption{Inferences under the CoT-LLM model against people for Exp. 1 (left) and Exp. 2 (right). Error bars depict standard deviation over the human and model responses.}
    \label{fig:e1e2-cot-scatter}
\end{figure*}

\begin{figure*}[h!]
    \centering
    \includegraphics[width=1.0\linewidth]{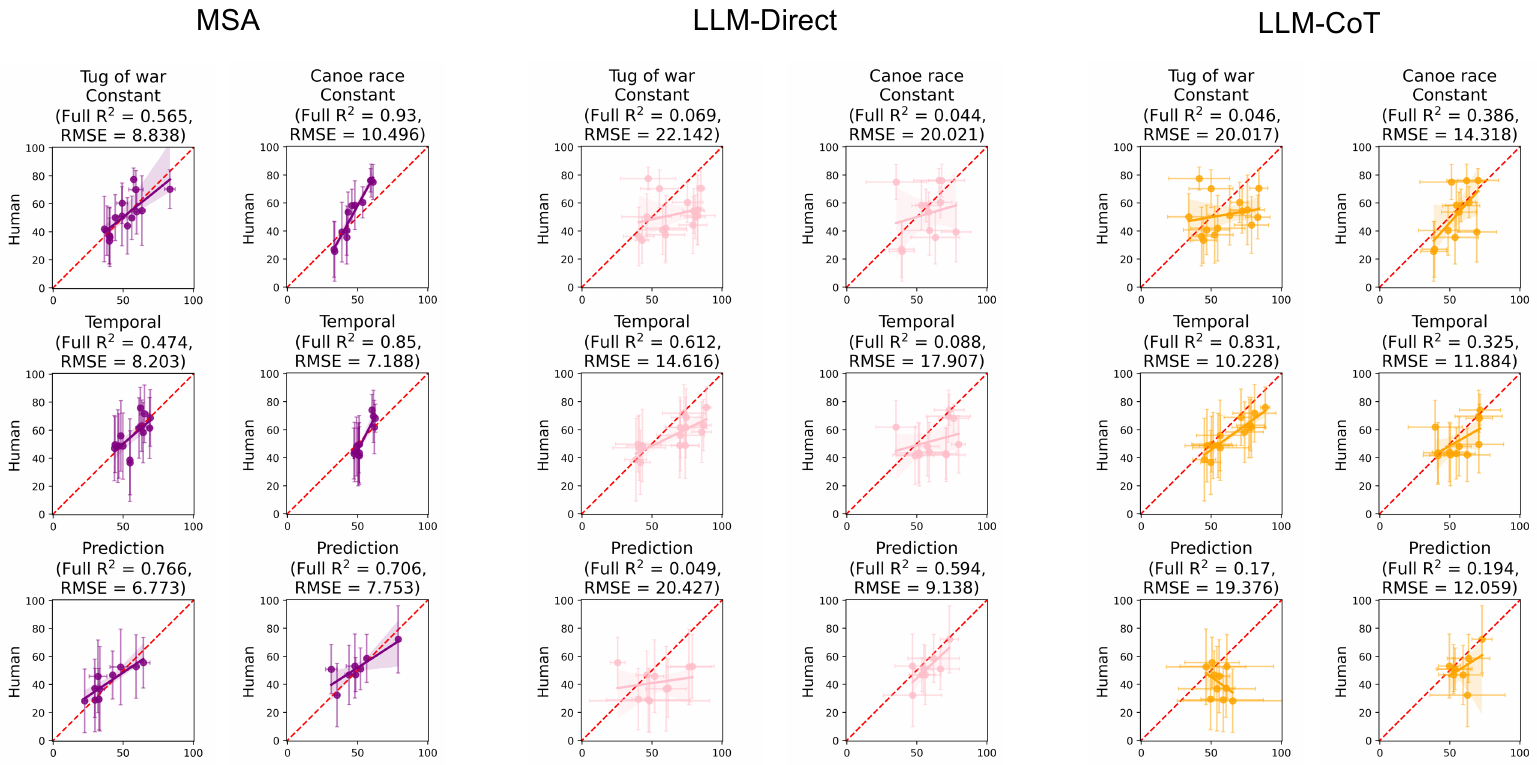}
    \caption{Inferences under MSA (left), Direct-LLM (middle), and CoT-LLM (right) against people for Exp. 3. Error bars depict standard deviation over the human and model responses.}
    \label{fig:e3-scatter}
\end{figure*}

\paragraph{Human-Model Correlations for All Models}
Below, we include the full set of scatterplots between the average human and average model responses for the three experiments. We depict additional gold model results for Exp. 1 and Exp. 2 in Figure~\ref{fig:e1e2-gold-scatter}, Direct-LLM in Figure~\ref{fig:e1e2-vanilla-scatter}, and CoT-LLM in Figure~\ref{fig:e1e2-cot-scatter}. We compare all model scatterplots on Exp. 3 in Figure~\ref{fig:e3-scatter}.  

\begin{figure*}[h!]
    \centering
    \includegraphics[width=\linewidth]{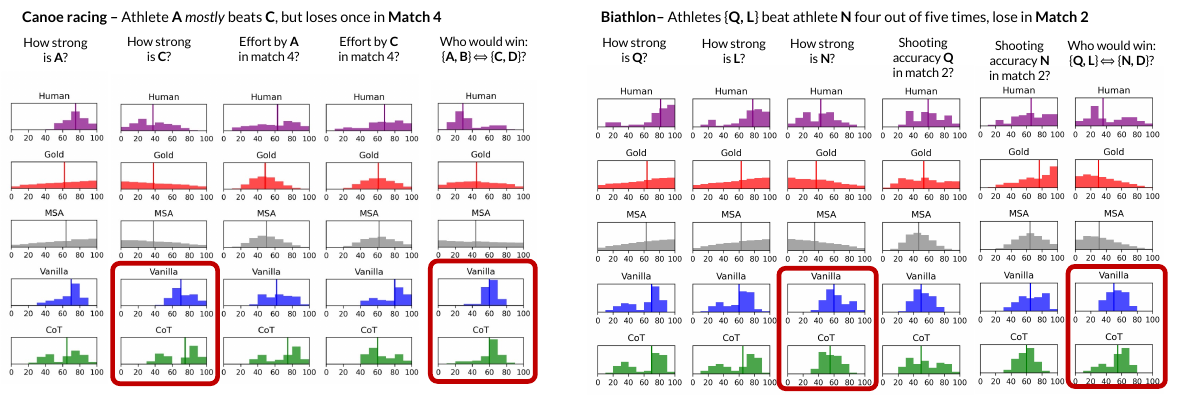}
    \vspace{-0.7cm}
    \caption{Illustrative examples from Experiments 1 and 2 highlighting one divergent pattern in inferences from LLM-only baselines, relative to human judgments (and normative Bayesian inferences in models synthesized by our MSA implementation). In the \textit{canoe racing scenario} (\textbf{left}), noisy evidence in the vignette suggests that athlete \textit{A} often appears in a winning pair of teammates that beat teams containing athlete \textit{C}, despite a single anomalous loss. Humans judge C to be largely weaker than average, but both LLM-baselines (red) switch to predicting C as particularly strong; and predict that C would now win on a team gainst A. Similarly, in the \textit{biathlon scenario} (\textbf{right}), a \textit{pair} of athletes (Q, L) frequently beats another (N), while losing anomalously once. LLM-baselines allocate more probability to the possibility that both Q and L are actually quite weak, largely believe N is stronger, and tend to predict that N will win against Q and L.}
    \label{fig:qualitative-histograms}
\end{figure*}

\paragraph{Human-Model Correlations for All Models} \autoref{fig:qualitative-histograms} also briefly summarizes qualitative error analysis patterns between Experiments 1 and 2, highlighting distinctions in LM-only baselines relative to human judgments in overall patterns of judgments (red) -- as well as distinctions between the MSA baselines in the \textit{qualitative} nature of the distributions of human judgments (LMs often appear to make peakier judgments relative to the symbolic model posteriors, which could be an artifact of the 5-sampling procedure).

\end{document}